\documentclass[output=paper,modfonts,nonflat]{langsci/langscibook}

\bibliography{localbibliography}
\usepackage{tabularx} 

\usepackage{./langsci/styles/langsci-optional}
\usepackage{./langsci/styles/langsci-gb4e}
\usepackage{./langsci/styles/langsci-lgr}

\renewbibmacro*{index:name}[5]{%
  \usebibmacro{index:entry}{#1}
    {\iffieldundef{usera}{}{\thefield{usera}\actualoperator}\mkbibindexname{#2}{#3}{#4}{#5}}}

\usepackage{epsfig,avm,gb4e}
\usepackage{amssymb} % jsilva: for \eth

\renewcommand{\textSigma}{$\Sigma$} % jsilva: not as pretty, but works

\title{Anaphoric Binding: \\an integrated overview} 
\author{%
Ant\'onio Branco\affiliation{University of Lisbon\\ 
NLX---Natural Language and Speech Group, Department of Informatics\\
Faculdade de Ci\^encias, Campo Grande, 1749-016 Lisboa, Portugal\\
\texttt{antonio.branco@di.fc.ul.pt}
}
}
\abstract{
The interpretation of anaphors depends on their antecedents as the semantic value that an anaphor eventually conveys is co-specified
by the value of its antecedent. Interestingly, when occurring 
in a given syntactic position, different anaphors may 
have different sets of admissible antecedents.
Such differences are the basis for the categorization of anaphoric expressions according to their anaphoric capacity, being important to determine what are the sets of admissible antecedents and how to represent and process this anaphoric capacity for each type of anaphor.

From an empirical perspective, these constraints stem
from what appears as quite cogent generalisations and exhibit a universal
character, given their cross linguistic validity. From a conceptual point of view, in turn, the relations among binding
constraints involve non-trivial cross symmetry, which lends them a modular nature
and provides further strength to the plausibility of their universal character.
This kind of anaphoric binding constraints appears thus as a most significant subset of natural language knowledge, usually referred to 
as binding theory.

This paper provides an integrated  overview of these constraints holding
on the pairing of nominal anaphors with their admissible antecedents
that are based on grammatical relations and structure.  Along 
with the increasing interest on neuro-symbolic approaches
to natural language,
this paper seeks to contribute
to revive the interest on this most intriguing research topic.}

\begin{document}

\maketitle
\rohead{Anaphoric Binding: an integrated overview} % jsilva: to get the running head without the linebreak
\section{Introduction} 

It is an inherent feature of natural languages that their expressions have 
the potential to convey a range of semantic values which their usage 
in context will constrain in such a way that some value is eventually 
circumscribed and expressed. This is exemplified in the minimal contrast below,
where out of its possible semantic values, the expression \textit{flying planes}
express one of such values in each context.

\is{flying planes}
\begin{exe}
\ex
\begin{xlist}
\ex{Flying planes are complex machines.}
\ex{Flying planes is a difficult task.}
\end{xlist}
\end{exe}

Like other natural language expressions, anaphors are also semantically polyvalent. They form however a class of expressions whose context
sensitiveness is rather peculiar in as much as for them to eventually express a semantic
value, more than being circumscribed from an intrinsic repertoire of potential
values, that value is co-specified by the semantic value of other expressions,
which are for this reason termed as their antecedents. This is exemplified in the contrasts below with the anaphor \textit{it}, that could
be continued at will. This anaphor inherently contributes the information that its denotation is singular and non human.
Yet in each different context, it eventually conveys a different semantic value as a result of that value being co-specified by the semantic value of a different antecedent (in italics),

\is{it examples}
\begin{exe}
\ex
\begin{xlist}
\ex{John pulled off \textit{the wheel}. It was heavy.}
\ex{Paul bought \textit{a computer}. It has a touch screen.}
\ex{Peter got \textit{a ticket that Paul wanted to buy}. It is for Saturday night.}
\ex{...}
\end{xlist}
\end{exe}

Adding to this semantic peculiarity, the dependency of anaphors with respect to their antecedents may exhibit the syntactic peculiarity of being a long-distance relation. This is illustrated with the set of contrasts below, that could be continued at will, where the anaphor \textit{her} and its antecedent anaphor \textit{Mary} can be separated by a string of words of arbitrary length.

\pagebreak
\is{it examples}
\begin{exe}
\ex
\begin{xlist}
\ex{Mary thinks that Peter saw her.}
\ex{Mary thinks that John knows that Peter saw her.}
\ex{Mary thinks that Paul believes that John thinks that Peter saw her.}
\ex{...}
\end{xlist}
\end{exe}

In large enough contexts, an anaphoric expression may have more
admissible antecedents than the antecedent that happen
to eventually co-specify its interpretation. This receives
a minimal example in the sentence below, where the anaphor \textit{herself}
has two admissible antecedents, out of which one will eventually end up being 
the selected antecedent given the respective utterance context (not represented below).

\begin{exe}
\ex \textit{Claire} described \textit{Joan} to herself.
\end{exe}

Interestingly, when occurring in a given syntactic position, different anaphoric expressions may 
have different sets of admissible antecedents. This is illustrated in the emblematic examples 
below, with two anaphoric expressions from \ili{English} --- \textit{herself} and \textit{her} --- 
occurring in the same position, each with different sets of admissible antecedents.

\begin{exe}
\ex Mary's friend knows that Paula's sister described Joan to herself / her.
\end{exe}

For the expression \textit{herself}, either \textit{Joan} or \textit{Paula's sister} is an admissible antecedent. 
For \textit{her}, its set of admissible antecedents includes instead \textit{Paula}, \textit{Mary's friend} and \textit{Mary}. Further examples with two anaphoric expressions occurring in the same position
and each with different sets of admissible antecedents are the following:

\begin{exe}
\ex Mary's friend knows that Paula's sister saw her / the little girl.
\end{exe}

For the expression \textit{the little girl}, either \textit{Paula} or \textit{Mary} is an admissible antecedent. 
For \textit{her}, its set of admissible antecedents includes additionally \textit{Mary's friend}.

Such differences in terms of sets of admissible antecedents are the basis for 
the partition of nominal anaphoric expressions into different groups according to their anaphoric
capacity. It has been an important topic of research to determine how many such groups or types of anaphoric expressions there are, what are the sets of admissible antecedents for each type, what expressions belong to which type in each language, and how to represent and process this anaphoric capacity.

The regularities emerging with this inquiry have been condensed in a handful of anaphoric 
binding constraints, or principles, which seek to capture the relative positioning of anaphors and their admissible antecedents in grammatical representations. 
From an empirical perspective, these constraints stem
from what appears as quite cogent generalisations and exhibit a universal
character, given their cross linguistic validity.%
\footnote{
\citep{branco:livro00}, \em{i.a.}
}
From a conceptual point of view, in turn, the relations among binding
constraints involve non-trivial cross symmetry, which lends them a modular nature
and provides further strength to the plausibility of their universal character.%
 \footnote{
\citep{branco:2005}.
}
Accordingly, anaphoric binding principles appear as one of
the most significant subsets of grammatical knowledge, usually termed as Binding Theory.

\centerline{$\star$}

This paper provides a condensed yet systematic and integrated overview of these grammatical constraints 
on anaphoric binding, that is of the grammatical constraints holding
on the pairing of nominal anaphors with their admissible antecedents. 
The integration into grammar of these anaphoric binding constraints in a formally sound and computationally tractable way, as well as of the appropriate semantic representation 
of anaphors are also covered in the present summary. On the basis of such overview, the ultimate goal of this paper is to provide an outlook into promising avenues for future
research on anaphoric binding and its modelling --- both from symbolic and neural perspectives.

\centerline{$\star$}

In the next section, Section~\ref{spec},
the empirical generalisations captured in the binding constraints are introduced, together
with the relevant auxiliary notions and parameterisation options.%
\footnote{
To support 
this presentation, the frameworks adopted are Head-Driven Phrase Structure Grammar \citep{polsag:hpsg94}, for syntax, and Minimal Recursion Semantics \citep{copesatke:mrs2005} and Underspecified Discourse Representation Theory \citep{frank:sem95},
for semantics.
The adoption of or transposition to other sufficiently expressive and well defined 
frameworks will be quite straightforward.}

The key ingredients for the integration of binding constraints into grammar
are discussed in Section \ref{sem}, and a detailed account of this integration is provided
in the following Section \ref{spec1} --- which is further illustrated with the support of
the working example in the Appendix.

Section \ref{discuss} is devoted to discuss how
the account of anaphoric binding presented in the previous sections ensures a neat interface of grammar with reference processing
systems, and thus supports a seamlessly articulation of binding constraints
with anaphora resolution.

In the penultimate section, Section \ref{reverse}, additional binding constraints are introduced,
that hold from the perspective of the antecedents, rather from the perspective of the anaphors,
together with the respective supporting empirical evidence.

The final Section \ref{outlook} is devoted to provide an outlook into promising avenues for future
research that may further enhance our understanding of and our coping with anaphoric binding and its modelling --- both symbolic and neural.

\section{Empirical generalisations \label{spec}}

\is{binding}
\is{anaphoric binding}
\is{binding principles}
\is{Principle A}
\is{Principle Z}
\is{Principle B}
\is{Principle C}
\is{reflexives}
\is{short-distance reflexives}
\is{long-distance reflexives}
\is{pronouns}
\is{non-pronouns}

Since the so called integrative approach to anaphora resolution was set up,%
\footnote{The integrative approach to anaphora resolution was set up in
\citep{carb:resol88, richluper:resol88, asher:resol89}, and its practical viability
was extensively checked out in \citep{lappin:pron94, mitkov:resol97}.
}
it is common wisdom that factors determining the antecedents of anaphors divide into 
filters, or hard constraints, and preferences, or soft constraints.  The former exclude
impossible antecedents and help to circumscribe the set of admissible antecedents; 
the latter interact to converge on the eventual antecedent among the admissible antecedents. 

So-called binding principles are a notorious subset of hard constraints on anaphora resolution:  
they capture generalisations concerning the constraints on the relative positioning of anaphors 
with respect to their admissible antecedents in the grammatical geometry of sentences.

We present below the definition of binding constraints,%
\footnote{
This is the approach
to Binding Theory proposed in \citep{polsag:binding92} and \citep[Chap.6]{polsag:hpsg94}, and subsequent developments in
\citep{xue:ziji94,branco:branch96,brancoMarrafa:subject97, manningSag:1999, wechsler:1999, 
koenig:equa99, branco:ldrefl99, richter:quant99, golde:diss99, branco:livro00, kiss:2001, branco:2002a, branco:2002b, branco:2002c} {\em i.a.}}
which resorts to a few auxiliary 
notions --- locality, o-command, o-binding ---, whose definition, in turn, are presented right afterwards.

There are four such constraints on the anaphoric capacity of nominals, named Principle A, Z, B and C. 
They induce a partition of the set of anaphors
into four classes. According to this partition, every nominal anaphor is of one of the
following anaphoric types: short-distance reflexive, long-distance reflexive, pronoun,
or non-pronoun.

The definition of each binding principle in (\ref{PrincipleA})-(\ref{PrincipleC}) is paired with an illustrative example with 
key grammatical contrasts
empirically supporting the respective generalisation. In particular, Principle A in (\ref{PrincipleA}) is paired with an example 
with the short-distance reflexive {\em himself},  Principle Z in (\ref{PrincipleZ}) is paired with the Portuguese long-distance reflexive {\em ele pr\'{o}prio}, Principle B in (\ref{PrincipleB}) with the pronoun {\em him}, and Principle C in (\ref{PrincipleC}) with the non pronoun {\em the boy}. These examples will be discussed below right after the definitions of the auxiliary notions have been presented.

%\begin{exe}
%\label{bindingPrinciples}
%\ex 
%\end{exe}
%
%\begin{equation}
%     \label{bindingPrinciples}
%\vspace*{-3 mm}
%\end{equation}

\begin{exe}
%\label{bindingPrinciples}
\ex
\label{PrincipleA}
{\textbf{Principle A:}} A locally o-commanded short-distance
reflexive must be \mbox{locally} o-bound.

\sn
{...{\em X}$_{x}$...[Lee$_{i}$'s friend]$_{j}$ thinks
[[Max$_{k}$'s brother]$_{l}$ likes %\linebreak
himself$_{*x/*i/*j/*k/l}$].}
\end{exe}

%\pagebreak

\begin{exe}
%\sn
\ex
\label{PrincipleZ}
{\textbf{Principle Z:}} An o-commanded long-distance reflexive must be
o-bound.

\sn
\gll ...{\em X}$_{x}$...[O amigo do Lee$_{i}$]$_{j}$ acha [que [o
irm\~{a}o do Max$_{k}$]$_{l}$ gosta dele pr\'{o}prio$_{*x/*i/j/*k/l}$]. (\ili{Portuguese})\\ \mbox{ }\mbox{ }\mbox{ }\mbox{ }\mbox{ }\mbox{ }\mbox{ }\mbox{ }\mbox{ }\mbox{ }\mbox{ }\mbox{ }the friend of.the Lee thinks \mbox{ }that \mbox{ }the brother of.the Max likes of.him self\\
\trans '...{\em X}$_{x}$...[Lee$_{i}$'s friend]$_{j}$ thinks [[Max$_{k}$'s brother]$_{l}$
likes him$_{*x/*i/j/*k}$ / \linebreak
himself$_{l}$].'
\end{exe}

\begin{exe}
%\sn
\ex
\label{PrincipleB}
{\textbf{Principle B:}} A pronoun must be locally o-free.

\sn
{...{\em X}$_{x}$...[Lee$_{i}$'s friend]$_{j}$ thinks [[Max$_{k}$'s brother]$_{l}$
likes %\linebreak
 him$_{x/i/j/k/*l}$].}
\end{exe}

\begin{exe}
%\sn
\ex
\label{PrincipleC}
{\textbf{Principle C:}} A non-pronoun must be o-free.

\sn {...{\em X}$_{x}$...[Lee$_{i}$'s friend]$_{j}$ thinks [[Max$_{k}$'s brother]$_{l}$
likes %\linebreak 
the boy$_{x/i/*j/k/*l}$].}
\end{exe}
\vspace{4 mm}

\subsection{Binding, coindexation, locality, command and crosslinguistic variation}\label{parameterisation}

The empirical generalisations presented above result from linguistic analysis supported by empirical
evidence of which the respective examples above are just a few key illustrative cases. These examples will be discussed
in detail in the next subsection, thus illustrating the analysis underlying the binding principles above.

The above definition of binding principles is rendered with the help of a few auxiliary
notions. For many of these auxiliary notions, their final value or definition is amenable to be
set according to a range of options: as briefly exemplified below, this parameterisation may be
driven by the particular language at stake, by the relevant predicator
selecting the anaphor, by the specific anaphoric form, etc.

These are the definitions of those auxiliary notions:

\is{o-binding}
{\bf Binding} {\em O-binding} is such that ``$x$ o-binds $y$ iff $x$ o-commands $y$ 
and $x$ and $y$ are coindexed" ({\em o-freeness} is non o-binding).%
\footnote{\citep[p.279]{polsag:hpsg94}.}

{\bf Coindexation} {\em Coindexation} is meant to represent an anaphoric link between the expressions 
with the same index. A starred index, in turn, indicates that the anaphoric link represented is not acceptable, as in the following examples:

\is{coindexation}
\begin{exe}
\ex
\begin{xlist}
\ex John_{i} said that Peter_{j} shaved himself_{*i/j}.
\ex John_{i} said that Peter_{j} shaved him_{i/*j}.
\end{xlist}
\end{exe}

Turning to example (\ref{PrincipleB}), for instance, {\em him}$_{k}$ and {\em Max}$_{k}$ are
coindexed with {\em k}, thus indicating their anaphoric binding and representing that {\em Max} is the antecedent
of {\em him}. The starred index {\em *l}, in turn, indicates that the coindexation between  {\em Max's brother}$_{l}$
and {\em him}$_{*l}$ is not felicitous, and thus that {\em Max's brother} is not an admissible 
antecedent of {\em him} in (\ref{PrincipleB}).

In the examples above, '...{\em X}$_{x}$...' represents a generic, extra-sentential
antecedent, available from the context.

Plural anaphors with so-called split antecedents, that has concomitantly
more than one antecedent, are represented with a sum of indexes as a subscript, as exemplified
below by {\em them} being interpreted as referring to John and Mary:%
\footnote{
When at least one of the antecedents in a split antecedent relation does not comply
with the relevant binding principle (and there is at least one that complies with it), 
the acceptability of that anaphoric link degrades. Apparently, the larger the number 
of antecedents that violate the binding constraint the less acceptable
is the anaphoric link: while both examples below are not fully acceptable,
two coindexations out of three, via $j$ and $k$,  in violation of the Principle B
render example b.~less acceptable than example a., which has one coindexation 
only, via $k$, in violation of that binding constraint  \citep[313]{seeley93}:

\is{split antecedents}
\begin{exe}
\ex
\begin{xlist}
\ex[?]{The doctor_{i} told the patient_{j} [that the nurse_{k} would protect them_{i+j+k} during the storm].}
\ex[??]{The doctor_{i} said [that the patient_{j} told the nurse_{k} about them_{i+j+k}].}
\end{xlist}
\end{exe}

As for plural reflexives, which in turn comply with Principle A, they accept split antecedents only in exempt 
positions --- on the notion of exemption, see Section \ref{Exemption}.}

\begin{exe}
\ex John_{i} told Mary_{j} that Kim talked about them_{i+j}.
\end{exe}

\is{local domain}
\is{binding local domain}
\textbf{Locality} The {\em local domain} of an anaphor results from
the partition of sentences and associated grammatical geometry into two
zones of greater or less proximity with respect to the anaphor.

Typically, the local domain coincides with the immediate selectional domain
of the predicator directly selecting the anaphor. In the following example, 
the local domain of {\em him} is explicitly marked within square brackets:

\begin{exe}
\ex John knows that [Peter described him].
\end{exe}

In the example in (\ref{PrincipleA}), for instance, {\em Max's brother} is immediately 
selected by {\em likes}, the predicator that immediately selects {\em himself}, while {\em Lee's friend}
is not. Hence, the first is in the local domain of {\em himself}, while the latter is not.

In some cases, there may be additional requirements that the local domain
is circumscribed by the first selecting predicator that happens to be finite, 
bears tense or indicative features, etc.%
\footnote{
Vd. \citep{manziniWexler:parameters87, kosterReuland:longdistance91, dal:bind93} for further details.
}
One such example can be the following:%
\footnote{
\citep[p.47]{manziniWexler:parameters87}.}

\begin{exe}
\ex
\begin{xlist}
\ex
\gll J\'{o}n$_{i}$ segir a$\eth$ [Maria$_{j}$ elskar sig$_{*i/j}$ ]. (\ili{Icelandic})\\
J\'{o}n says-\textsc{ind} that \mbox{ }Maria loves-\textsc{ind} himself\\
\trans 'J\'{o}n$_{i}$ says that [Maria$_{j}$ loves himself$_{*i}$/herself$_{j}$].'

\ex
\gll [J\'{o}n$_{i}$ segir a$\eth$ Maria$_{j}$ elski sig$_{i/j}$ ].\\
J\'{o}n says-\textsc{ind} that Maria loves-\textsc{subj} himself\\
\trans '[J\'{o}n$_{i}$ says that Maria$_{j}$ loves himself$_{i}$/herself$_{j}$].'
\end{xlist}
\end{exe}

In the first sentence above, the verb in the embedded clause is
Indicative and the local domain of its Direct Object is circumscribed
to this clause as the reflexive cannot have the Subject of the upwards
clause as its antecedent. The second sentence is identical to the first
one except that the mood of the embedded verb is now Subjunctive. This leads
to a change in the local domain of the reflexive: it can now have also the upwards
Subject as its antecedent, thus revealing that its local domain is determined
by the first selecting verb in the Indicative, which happens now to be the
verb of the upwards clause.

In some other languages, there are anaphors whose local domain
is the immediate selectional domain not of the directly selecting predicator
but of the immediately upwards predicator,  irrespective of the
inflectional features of the directly or indirectly selecting predicators. 
This seems to be the case of the \ili{Greek} {\it o idhios}:%
\footnote{
Alexis Dimitriadis p.c. See also \citep{iatridou:86, varlokostaHornstein:93}. }

\enlargethispage*{2mm}

\begin{exe}
\ex
\gll O Yannis$_{i}$ ipe stin Maria [oti o Costas$_{j}$ pistevi [oti o Vasilis$_{k}$ aghapa ton
idhio$_{??i/j/*k}$]]. (\ili{Greek})
\\ 
the Yannis told the Maria \mbox{ }that the Costas believes \mbox{ }that the Vasilis loves the same.\\
\trans 'Yannis$_{i}$ told Maria that [Costas$_{j}$ believes that [Vasilis$_{k}$ loves
him$_{??i/j/*k}$]].'
\end{exe}

Languages shows diversity concerning which of these options are materialized
and which grammatical and lexical means are brought to bear.%
\footnote{
\citep{dimitriadisDatabase:2005}.
}
Additionally, not all languages have anaphors of every one of the anaphoric types:
For instance, \ili{English} is not known to have long-distance reflexives.

\is{command}
\is{o-command}
\is{obliqueness}
{\bf Command} {\em O-command} is a partial order defined on the basis of the obliqueness hierarchies
of grammatical functions, possibly embedded in each other along the relation of
subcategorisation:``Y o-commands Z just in case either Y is less
oblique than Z; or Y {\mbox o-commands} some X that subcategorises for Z; or Y
o-commands some X that is a projection of Z".%
\footnote{\citep[p.279]{polsag:hpsg94}.}

The grammatical function Subject is less oblique than the Direct Object, the Direct
Object is less oblique than the Indirect Object, etc., thus establishing a so-called obliqueness hierarchy. 
The obliqueness hierarchy
of grammatical functions is represented in the value of ARG-ST feature in as much as 
the arguments are ordered from those whose grammatical function is less oblique to those 
whose function is more oblique.
As discussed in detail in Section \ref{spec1} and in connection with the working example 
in the Appendix, ARG-ST feature value plays a crucial role in the formalization
and explicit integration of binding principles into grammar.\footnote{
For further discussion of the notion of obliqueness of grammatical functions as well as further references on
this topic, see \citep[Sec.5.2]{polsag:hpsg87}.}

Accordingly, the Subject o-commands the Direct Object, the Direct
Object o-commands the Indirect Object, etc.;
and in a multi-clausal sentence,
the arguments in the upwards clauses o-command the arguments in the
successively embedded clauses.

\begin{exe}
\ex $[[$John's friend$]$ said that $[[$Peter's brother$]$ presented [Martin's cousin] to him$]]$.
\end{exe}

In the example above, {\em John's friend} o-commands {\em Peter's brother}, {\em Peter}, 
{\em Martin's cousin}, {\em Martin} and {\em him}. 
{\em Peter's brother} locally o-commands {\em Martin's cousin} and {\em him},
and (non-locally) o-commands {\em Martin}. Neither {\em John}, {\em Peter}, {\em Martin} nor {\em him} is o-commanding
any nominal in this example.

In the example of (\ref{PrincipleZ}), for instance, the Portuguese long-distance reflexive {\em ele pr\'{o}prio},
which is the Object of the embedded clause, is o-commanded by {\em o irm\~{a}o do Max} (Max's brother),
the Subject of that clause, and by {\em o amigo do Lee} (Lee's friend), the Subject of the upwards clause.
{\em Lee} and {\em Max}, in  turn, do not o-command this reflexive in this example.

\subsection{Binding principles}\label{principles}

With the definition of the auxiliary notions above in place, the definition 
of binding principles in (\ref{PrincipleA})-(\ref{PrincipleC}) is now complete and it is possible
to appreciate how the respective examples instantiate them.

{\bf Principle A} The example in (\ref{PrincipleA}) shows that the anaphoric capacity of {\em himself} complies with 
the anaphoric discipline captured by Principle A: if it is locally o-commanded, it has to be locally o-bound, 
i.e. only locally o-commanders can be its admissible antecedents if it happens to be locally o-commanded.

{\em Max's brother} is in the local domain
of {\em himself} because it is immediately selected by the predicator {\em likes} which also
immediately selects {\em himself}. Moreover, {\em Max's brother}, in a Subject position, 
o-commands {\em himself}, in an Object position. {\em Max's brother}
is thus a local o-commander of {\em himself}, and hence it is an admissible antecedent
of {\em himself}. 

The other nominals in this example are not local o-commanders of {\em himself}: both
{\em Lee} and {\em Lee's friend} are selected by the main clause predicator {\em thinks},
not by the predicator {\em likes} which is immediately selecting {\em himself} and are thus not 
in its local domain; {\em Max} in turn, given it is embedded inside the local Subject, it is not immediately selected 
by the predicator {\em likes} that is immediately selecting {\em himself}. Hence, none of
the nominals in the sentence other than {\em Max's brother} happen to be local o-commanders of {\em himself} 
and thus are not one of its admissible antecedents.

Also any other antecedent candidate eventually available in the extra-sentential context is not
a local o-commander of {\em himself} and thus it is not one of its admissible antecedents.

Given the anaphoric capacity of {\em himself} complies with  the anaphoric discipline captured 
by Principle A, it belongs to the class of short-distance reflexives.

In connection with Principle A, it is also worth signalling that it is not the case that only Subjects
can be local o-commanders of short-distance reflexives, as illustrated below.

\begin{exe}
\ex
\label{objectsSDR}
\begin{xlist}

\ex Peter$_{i}$ didn't talk to John$_{j}$ about himself$_{i/j}$.
\label{objectsSDRunmarked}
 
\ex About himself$_{i/j}$, Peter$_{i}$ didn't talk to John$_{j}$.
\label{objectsSDRtopicalization}
\end{xlist}

\end{exe}

In the examples in (\ref{objectsSDR}), {\em John} and {\em himself} are in the same local domain.
Moreover, {\em John}, in the Object position, is less oblique than {\em himself}, in the Indirect Object position. Hence, {\em John}
is a local o-commander of the reflexive and qualifies as one of its admissible
antecedents, together with {\em Peter}, in the Subject position.

\is{c-command}
The absence of contrast between 
(\ref{objectsSDRunmarked}) and (\ref{objectsSDRtopicalization}) is a central piece of evidence that 
the command relation for anaphoric binding is based on the obliqueness hierarchy of grammatical functions (o-command) 
rather than on a configurational hierarchy based on surface syntactic structure (c-command).%
\footnote{Binding principles based on o-command, rather than on c-command as proposed in \citep{chom:bind80} and \citep{chom:knowledge86}, 
is a hallmark of the analysis in \citep{polsag:binding92}.

The analysis based on c-command incorrectly predicts
that anaphoric links like those in (\ref{objectsSDRtopicalization}) would not be acceptable, because the
admissible antecedents of {\em himself} do not c-command it (it is {\em himself} that c-commands them instead).

The analysis based on c-command also incorrectly predicts
that the anaphoric link between {\em himself} and {\em John} in (\ref{objectsSDRunmarked}) would not be acceptable, because
{\em John} does not c-command {\em himself}.

For a detailed discussion, see \citep[Chap.6]{polsag:hpsg94}.
}

{\bf Principle Z} The example in (\ref{PrincipleZ}) shows that the anaphoric capacity of 
the Portuguese nominal {\em ele pr\'{o}prio} complies with 
the anaphoric discipline captured by Principle Z: if it is o-commanded, it has to be o-bound, 
i.e. only o-commanders can be its admissible antecedents if it happens to be o-commanded.

{\em ele pr\'{o}prio} is (locally) o-commanded by {\em o irm\~{a}o do Max} because both are selected 
by the predicator {\em gosta} and {\em o irm\~{a}o do Max} is less oblique than {\em ele pr\'{o}prio}.
{\em ele pr\'{o}prio} is also o-commanded
by {\em o amigo do Lee} because {\em o amigo do Lee} is selected by the predicator of the upwards
clause {\em acha}, which selects the embedded clause whose predicator selects
{\em ele pr\'{o}prio}, and {\em o amigo do Lee} is thus less oblique than {\em ele pr\'{o}prio}
in the composite obliqueness hierarchy. 

The other nominals in this example are not o-commanders of {\em ele pr\'{o}prio}: both
{\em Lee} and {\em Max} are embedded inside arguments of the relevant predicators {\em acha}
and {\em gosta} but are not arguments of them. Hence, none of
the nominals in the sentence other than {\em o amigo do Lee} and  {\em o irm\~{a}o do Max} happen 
to be o-commanders of {\em ele pr\'{o}prio} and thus are not one of its admissible antecedents.

Also any other antecedent candidate eventually available in the extra-sentential context is not
an o-commander of {\em ele pr\'{o}prio} and thus it is not one of its admissible antecedents.

Given the anaphoric capacity of {\em ele pr\'{o}prio} complies with  the anaphoric discipline captured 
by Principle Z, it belongs to the class of long-distance reflexives.

{\bf Principle B} The example in (\ref{PrincipleB}) shows that the anaphoric capacity of {\em him} complies with 
the anaphoric discipline captured by Principle B: it has to be locally o-free, i.e. its local o-commanders cannot
be its admissible antecedents.

In that example, {\em Max's brother} is the only local
o-commander of {\em him} because {\em Max's brother} is the only argument of the predicator {\em likes} other than {\em him},
and is less oblique than {\em him}.

The other nominals in this example are not local o-commanders of {\em him}: neither {\em Lee's friend},
{\em Lee} or {\em Max} are immediately selected by {\em likes}.

Also any other antecedent candidate eventually available in the extra-sentential context is not
a local o-commander of {\em him}.

Hence, in this example all antecedent candidates, sentential and non sentential, are admissible 
antecedents of {\em him} except {\em Max's brother}.

Given the anaphoric capacity of {\em him} complies with  the anaphoric discipline captured 
by Principle B, it belongs to the class of pronouns.

{\bf Principle C} The example in (\ref{PrincipleC}) shows that the anaphoric capacity of {\em the boy} complies with 
the anaphoric discipline captured by Principle C: it has to be o-free, i.e. its o-commanders are not admissible antecedents. 

In that example, {\em Lee's friend} and {\em Max's brother} are the only o-commanders of {\em the boy}: 
{\em Lee's friend} is selected by the predicator of the upwards
clause {\em likes}, which selects the embedded clause whose predicator selects
{\em the boy}; {\em Max's brother}, in turn, is the only argument immediately selected by 
the predicator {\em likes} other than {\em the boy};
and both {\em Lee's friend} and {\em Max's brother} are less oblique than {\em the boy}.

The other nominals in this example are not o-commanders of {\em the boy}: neither {\em Lee} or {\em Max} 
are immediately selected by {\em thinks} or {\em likes}.

Also any other antecedent candidate eventually available in the extra-sentential context is not
an o-commander of {\em the boy}.

Hence, in this example all antecedent candidates, sentential and non sentential, are admissible antecedents of {\em the boy} except {\em Lee's friend} and {\em Max's brother}.

Given the anaphoric capacity of {\em the boy} complies with  the anaphoric discipline captured 
by Principle C, it belongs to the class of non pronouns.

\is{o-bottom}
\is{local domain reshuffling}
\is{binding exemption}
\subsection{O-bottom positions: reshuffling and exemption}\label{Exemption}

For the interpretation of an anaphor to be accomplished,
an antecedent has to be found for it. Such an antecedent is to be picked from
the set of its o-commanders, if the anaphor is a long-distance reflexive, 
or from the set of its local o-commanders, if it is a short-distance reflexive.

This requirement may not be satisfied in some specific cases, namely
when the reflexive occurs in a syntactic position such that it is the
least element of its \mbox{o-command} order, in an o-bottom position for short. In such
circumstances, it has no \mbox{o-commander} (other than itself, if the o-command relation
is formally defined as a reflexive relation) to qualify as its antecedent.

That is the motivation for the conditional formulation of Principles A and Z, 
in (\ref{PrincipleA}) and (\ref{PrincipleZ}) respectively:
a (short/) long-distance reflexive has to be (locally/) o-bound if it is (locally/) o-commanded.
In case it is not (locally/) o-commanded, there is no imposition concerning
their admissible antecedents following from Principles A and Z.

\textbf{Reshuffling} As a consequence, in some cases, the binding domain for the reflexive 
which happens to be the least element of its local obliqueness order may be reshuffled, being reset as 
containing the o-commanders of the reflexive in the domain 
circumscribed by the immediately upwards predicator.%
\footnote{
\citep{brancoHpsg:2005}.
}
One such case for a nominal domain can be found in the following example:%
\footnote{Tibor Kiss p.c., which is a development with regards to his 
data in \citep{kiss:2001}.}

\begin{exe}
\ex
\begin{xlist}
\ex
\gll Gernot$_{i}$ dachte, dass Hans$_{j}$ dem Ulrich$_{k}$ [Marias$_{l}$ Bild von
sich$_{*i/*j/*k/l}$] \"{u}berreichte. (\ili{German})\\
Gernot thought that Hans the Ulrich \mbox{ }Maria's picture of self gave\\
\trans 'Gernot$_{i}$ thought that  Hans$_{j}$ gave Ulrich$_{k}$ [Maria$_{l}$'s picture of \linebreak himself$_{*i/*j/*k}$/herself$_{l}$].'
 
\ex
\gll Gernot$_{i}$ dachte, dass [Hans$_{j}$ dem Ulrich$_{k}$ ein Bild von sich$_{*i/j/k}$ \"{u}berreichte].\\
Gernot thought that \mbox{ }Hans the Ulrich a picture of self gave\\
\trans 'Gernot$_{i}$ thought that  [Hans$_{j}$ gave Ulrich$_{k}$ [a picture of \linebreak himself$_{*i/j/k}$]].'
\end{xlist}
\end{exe}

In the first sentence above, the short-distance reflexive is 
locally \mbox{o-commanded} by {\em Maria} and only this nominal can be its
antecedent. In the second sentence, the reflexive is the first element
in its local obliqueness hierarchy and its admissible 
antecedents, which form now its local domain, are the nominals in the obliqueness
hierarchy of the immediately upwards predicator.

The null subject in languages like \ili{Portuguese} is another example of a short-distance 
reflexive that is in an o-bottom position and whose local domain is reshuffled:%
\footnote{
\citep{brancoNullSubject:2007}.
}

\is{null subject}
\begin{exe}
\ex
		\gll O m\'{e}dico_{i} disse-me [que [o director do Pedro_{j}]_{k} ainda n\~{a}o reparou [que $\emptyset_{*i/*j/k}$ cometeu um erro]]. (Portuguese)\\
		the doctor told-me \mbox{ }that \mbox{ }the director of.the Pedro yet not noticed \mbox{ }that { } made a mistake.\\
		\trans 'The doctor_{i} told me [that [Pedro_{j}'s director]_{k} didn't notice yet [that he_{*i/*j/k} made a mistake]].'

\end{exe}

In the example above, as the null reflexive is in an o-bottom position, its local domain gets reshuffled to include the immediately
upwards o-commander {\em Pedro's director}. Once it is thus o-commanded, in accordance do Principle A, the null reflexive cannot take other nominal in the sentence,
viz. {\em the doctor} or {\em Pedro}, as its admissible antecedent given none of these o-commands it.

%When there is no chance for such resetting, the reflexive turns out not to be interpretable:%
%%
%\footnote{\citep[p.173]{kiss:2003}.}
%
%\begin{exe}
%\ex
%\gll *Sich$_{i}$ friert. (German)\\
%\hspace*{2 mm}Self is\_cold\\
%\end{exe}

\textbf{Exemption} In some other cases, this resetting of the binding domain is not 
available. In such cases, the reflexive is in the bottom of its local obliqueness
order and is observed to be exempt of its typical binding regime: the reflexive may take antecedents 
that are not its o-commanders or that are
outside of its local or immediately upward domains,\footnote{
\citep[p.263]{polsag:hpsg94}.}
as illustrated in the following example:\footnote{
\citep[p.73]{golde:diss99}.
}

\begin{exe}
\ex Mary_{i} thought the artist had done a bad job, and was sorry
that her parents
came all the way to Columbus just to see the portrait of herself_{i}.
\end{exe}

In an exempt position, a reflexive can even have so-called
split antecedents, as illustrated in the following example
with a short-distance reflexive:%
\footnote{
\citep[p.42]{zribi:pview89}.
}

\begin{exe}
\ex Mary$_{i}$ eventually convinced her sister Susan$_{j}$ that John had better 
pay visits to everybody except themselves$_{i+j}$.
\end{exe}

That is an option not available for reflexives in non exempt positions:

\begin{exe}
\ex Mary$_{i}$ described$_{j}$ John to themselves$_{*(i+j)}$.
\end{exe}

Some long-distance reflexives may also be exempt from their binding constraint if
they occur in the bottom of their o-command relation. In such cases, they
can have an antecedent in the previous discourse sentences or
in the context, or a deictic use, as illustrated in the following example:

\begin{exe}
\ex
\gll [O Pedro e o Nuno]$_{i}$ tamb\'{e}m conheceram ontem a Ana. Eles pr\'{o}prios$_{i}$
ficaram logo a gostar muito dela. (\ili{Portuguese})\\ 
the Pedro and the Nuno also met yesterday the Ana. They {\em pr\'{o}prios} stayed immediately to
liking much of.her\\
\trans '[Pedro and Nuno]$_{i}$ also met Ana yesterday. They$_{i}$ liked her very much right
away.'
\end{exe}

Such options are not available in non exempt positions:%
\footnote{
For further details, vd. \citep{branco:ldrefl99}.
}

\begin{exe}
\label{portugueseLDreflexive}
\ex
\gll  A Ana tamb\'{e}m conheceu ontem [o Pedro e o Nuno]$_{i}$. Ela ficou logo a gostar muito
deles pr\'{o}prios$_{*i}$. (Portuguese)\\ 
The Ana also met yesterday \mbox{ }the Pedro and the Nuno. She stayed immediately to liking much of.them
{\em pr\'{o}prios}\\
\trans 'Ana also met [Pedro and Nuno]$_{i}$ yesterday. She liked them$_{*i}$ very much right
away.'
\end{exe}

Admittedly, an overarching interpretability condition is in force in natural languages
requiring the ``meaningful'' anchoring of anaphors to antecedents. Besides this general 
requirement, anaphors are concomitantly ruled by specific constraints concerning their 
particular anaphoric capacity, including the sentence-level constraints in (\ref{PrincipleA})-(\ref{PrincipleC}), i.e. the binding principles.

When reflexives are in o-bottom positions, an o-commander (other than the reflexive itself) may not be
available to function as antecedent and anchor their interpretation. Hence, such specific
binding constraints, viz. Principle A and Z, cannot be satisfied in a ``meaningful" way and the general
interpretability requirement may supervene them. As a consequence, in cases displaying so-called
exemption from binding constraints, o-bottom reflexives appear to escape their specific binding regime
to comply with such general requirement and its interpretability be rescued. 

The anaphoric links of exempt reflexives have been observed to be
governed by a range of non sentential factors (from discourse, dialogue, non linguistic context,
etc.), not being determined by the sentence-level binding principles in (\ref{PrincipleA})-(\ref{PrincipleC}).%
\footnote{For further details, vd. \citep{kuno:func87, zribi:pview89, golde:diss99} among others.
}

\is{subject orientedness}
\is{alternations}
\subsection{O-command: alternations and subject-orientedness}\label{ocommand}

\is{alternations}
\textbf{Alternations} In languages like \ili{English}, the o-command order can be established over the
obliqueness hierarchies of active and passive sentences alike:%
\footnote{
\citep{Jackendoff72a-u, polsag:hpsg94}.}

\is{passive}
\begin{exe}
\ex
\begin{xlist}
\ex[]{John$_{i}$ shaved himself$_{i}$.}
\label{acti}
\ex[]{John$_{i}$ was shaved by himself$_{i}$.}
\label{passi}
\end{xlist}
\end{exe}

The obliqueness hierarchy of grammatical functions is represented in ARG-ST 
and in both ARG-ST values of (\ref{acti}) and of (\ref{passi}), {\em John}
appears as the Subject and qualifies as a local o-commander of {\em himself},
and thus as an admissible antecedent of this reflexive.

In some other languages, only the obliqueness hierarchy of a given syntactic alternation 
is available to support
the o-command order relevant for binding constraints in both alternations.

This is the
case, for example, of the alternation active/objective voice in \ili{Toba Batak}. 
In this language, a reflexive in Object position of an active voice sentence
can have the Subject as its antecedent, but not vice-versa:%
\footnote{
\citep[p.72]{manningSag:1999}.}

\begin{exe}
\ex
\begin{xlist}
\ex
\gll mang-ida diri-na$_{i}$ si John$_{i}$. (\ili{Toba Batak})\\ 
[{\sc active}-saw himself\textsubscript{\textsc{Object}}]\textsubscript{VP} {\sc pm} John\textsubscript{\textsc{Subject}} \\
\trans 'John$_{i}$ saw himself$_{i}$.'

\ex
\gll mang-ida si John$_{i}$ diri-na$_{*i}$.\\ 
[{\sc active}-saw {\sc pm} John\textsubscript{\textsc{Object}}]\textsubscript{VP} himself\textsubscript{\textsc{Subject}} \\
\end{xlist}
\end{exe}

Taking the objective voice paraphrase corresponding to the active sentence
above, the binding pattern is inverted: a reflexive in Subject position can have
the Object as its antecedent, but not vice-versa, thus revealing that
the obliqueness hierarchy relevant for the verification of its binding
constraint remains the hierarchy of the corresponding active voice
sentence above:

\begin{exe}
\ex
\begin{xlist}
\ex
\gll di-ida diri-na$_{*i}$ si John$_{i}$.\\ 
[{\sc objective}-saw himself\textsubscript{\textsc{Object}}]\textsubscript{VP} {\sc pm} John\textsubscript{\textsc{Subject}} \\

\ex
\gll di-ida si John$_{i}$ diri-na$_{i}$.\\ 
[{\sc objective}-saw {\sc pm} John\textsubscript{\textsc{Object}}]\textsubscript{VP} himself\textsubscript{\textsc{Subject}} \\
\trans 'John$_{i}$ saw himself$_{i}$.'
\end{xlist}
\end{exe}

\is{subject-orientedness}
\textbf{Subject-orientedness} O-command may take the shape of a linear or non linear order
depending on the specific obliqueness hierarchy upon which it is 
realised.

In a language like \ili{English}, the arguments in the
subcategorisation frame of a predicator are typically arranged in a linear obliqueness
hierarchy.

In some other languages, the obliqueness hierarchy upon which the o-command order
is based may happen to be non linear: in the subcategorisation frame of a
predicator, the Subject is less oblique than any other argument while the remaining arguments
are not comparable to each other under the obliqueness relation. As a consequence, in a clause,
a short-distance reflexive with an Indirect Object grammatical function, for instance, may only
have the Subject as its antecedent, its only local o-commander.%
\footnote{For a thorough argument and further evidence motivated
independently of binding facts see \citep{branco:branch96, brancoMarrafa:subject97, branco:livro00}. In 
some languages, there can be an additional requirement that the Subject be
animate to qualify as a commander to certain anaphors. On this, 
see \citep{huangTang:longdistance91, xue:ziji94} about Chinese {\em ziji}, among others.}

This Subject-orientedness effect induced on the anaphoric capacity of reflexives
by the non linearity of the o-command relation can be observed in contrasts like
the following:%
\footnote{
Lars Hellan p.c. See also \citep[p.67]{hellan:book88}.}

\is{subject-oriented anaphor}
\begin{exe}
\ex
\begin{xlist}
\ex
\gll Lars$_{i}$ fortalte Jon$_{j}$ om seg selv$_{i/*j}$. (\ili{Norwegian})
\\ 
Lars told Jon about self {\em selv} \\
\trans 'Lars$_{i}$ told Jon$_{j}$ about himself$_{i/*j}$.'

\ex
\gll Lars$_{i}$ fortalte Jon$_{j}$ om ham selv$_{*i/j}$.\\
Lars told Jon about him {\em selv}\\
\trans 'Lars$_{i}$ told Jon$_{j}$ about him$_{*i/j}$.'
\end{xlist}
\end{exe}

In the first sentence above, the reflexive cannot have the
Direct Object as its antecedent given that the Subject is its only
local  o-commander in the non linear obliqueness hierarchy.
In the second sentence, under the same circumstances, a pronoun
presents the symmetric pattern: it can have any co-argument as
its antecedent except the Subject, its sole local o-commander.%
\footnote{
For an analysis of the Subject-orientedness of French {\em se} resorting
to a notion of s-command, see \citep{abeille:depend98, abeille:composition98}.
}

\section{Binding Constraints at the Syntax-Semantics Interface\label{sem}}

Like other sorts of constraints on semantic composition, binding
constraints impose grammatical conditions on the interpretation of certain expressions
--- anaphors, in the present case --- based on syntactic geometry.%
\footnote{
For a discussion of proposals in the literature that have tried to root binding principles on non-grammatical,
cognitive search optimisation mechanisms, and their pitfalls, see %\citep{branco:2000,branco:2003,branco:2004}
\citep{branco:2004,branco:2003,branco:2000}.
}
This should not
be seen, however, as implying that they
express grammaticality requirements. By replacing, for instance, a pronoun 
by a reflexive in a sentence, we are not turning a grammatical construction into 
an ungrammatical one, even if we assign to the reflexive the 
antecedent adequately selected for the pronoun. In that case, we 
are just asking the hearer to try to assign to that sentence a 
meaning that it cannot express, in the same way as what would 
happen if we asked someone whether he could interpret {\it The red book 
is on the white table} as describing a situation where a white 
book is on a red table. 

In this example, given how they happen to be syntactically related, 
the semantic values of {\it red} and {\it table} cannot be composed in a way 
that this sentence could be used to describe a situation 
concerning a red table, rather than a white table. 

Likewise, if we 
take the sentence {\it John thinks Peter shaved him}, given how they happen to be 
syntactically related, the semantic values of {\it Peter} and {\it him} cannot 
be composed in a way that this sentence could be used to describe a 
situation where John thinks that Peter shaved himself, i.e.\ Peter, rather than
a situation where John thinks that Peter shaved other people, e.g.\ Paul,
Bill, etc., or even John himself. 

The 
basic difference between these two cases is that, while in the 
first the composition of the semantic contributions of {\it white} and 
{\it table} (for the interpretation of their NP {\it white table}) is constrained by local
syntactic geometry, in the
latter the composition of the semantic contributions of {\it John} 
and {\it him} (for the interpretation of the NP {\it him}) is 
constrained by non-local syntactic geometry.

These grammatical constraints on anaphoric binding should thus be taken as conditions on semantic interpretation
given that they delimit (non-local) aspects of meaning composition,
rather than aspects of syntactic wellformedness.%
\footnote{
This approach is in line with \citep{gawron:anaph90}, and departs from other approaches where binding constraints have been viewed as 
wellformedness conditions, thus belonging to the
realm of Syntax: ``[they] capture  the distribution of pronouns and reflexives"
\citep[p.657]{rein:refl93}.}

These considerations leads one to acknowledge that, semantically, an anaphor 
should be specified in the lexicon as a function whose argument is
a suitable representation of the context --- providing a
semantic representation of the NPs available in the discourse vicinity ---, 
and delivers both an 
update of its anaphoric potential~--- which is instantiated as 
the set of its grammatically admissible antecedents --- and an update of the 
context, against which other NPs are interpreted.%
\footnote{
\citep{brancoDaarc:1998,brancoColing:2000,branco:2002a}.
}
Naturally, all in all, 
there will be four of
such functions available to be lexically associated to anaphors,
each corresponding to one of the different four classes of anaphors,
in accordance with the four binding constraints A,~Z,~B~or~C.%
\footnote{
This is in line with~\citep{johnson:disc90} concerning the 
processing of the semantics of nominals, and also 
the spirit (but by no means the letter) of the dynamic semantics framework~---~vd.~\ \citep{chi:dyn95} and \citep{stal:context98} {\em i.a.}
}

\subsection{Semantic patterns}\label{semanticPatterns}

\is{reference marker}
For an anaphoric nominal {\em w}, the relevant input context may be represented
in the form of a set of  three lists of reference markers,%
\footnote{
See \citep{Karttunen1976, Kamp1981, Heim1982, Seuren1985, kamp:drt93} for the notion of reference marker.
}
{\bf A}, {\bf Z} and
{\bf U}.  List {\bf A} contains the reference markers of the local
\mbox{o-command} order where {\it w} is included, ordered according to their relative grammatical
obliqueness; {\bf Z} contains  the markers of the (local and non local) \mbox{o-command} order where {\it w}
is included, i.e.\ reference markers organised in a possibly multi-clausal o-command relation, 
based upon successively embedded clausal obliqueness hierarchies;
and {\bf U} is the list
of all reference markers in the  discourse context, possibly including those not linguistically
introduced.

The updating of the context by an anaphoric nominal {\it w} may be seen
as consisting simply in the
incrementing of the representation of the context, with 
a copy of the reference marker of {\it w} being added to
the three lists above.

The updating of the anaphoric potential of {\it w}, in turn, delivers a representation
of the contextualised anaphoric potential of {\it w} in the form of the
list of reference markers of its admissible antecedents. This list
results from the binding constraint associated to {\it w} being applied  to
the relevant representation of the context of {\it w}.

Given this setup, the algorithmic verification of 
binding constraints consists of a few simple operations, and their
grammatical specification will consist thus in stating each such sequence of
operations in terms of the grammar description formalism. 

If the nominal {\it w} is a short-distance
reflexive, its semantic  representation is updated with {\bf A'}, where {\bf A'}
contains the reference  markers of the \mbox{o-commanders} of {\it w} in {\bf A}. 

If {\it w} is a
long-distance  reflexive, its semantic representation includes {\bf Z'}, such that {\bf Z'}
contains the \mbox{o-commanders} of {\it w} in {\bf Z}. 

If {\it w} is a pronoun, its semantics 
should include the list of its non-local \linebreak o-commanders,
that is the list
{\bf B}={\bf U}$\backslash$({\bf A'}$\cup$[r-mark$_{w}$])
is encoded into its semantic representation, where r-mark_{w} is the reference marker
of {\it w}.

Finally if {\it w} 
is a non-pronoun, its updated semantics keeps a copy of 
list \linebreak {\bf C}={\bf U}$\backslash$({\bf Z'}$\cup$[r-mark$_{w}$]),
which contains the non-o-commanders of {\it w}.

\subsection{Binding principles and other constraints for anaphora resolution}

These lists {\bf A'}, {\bf Z'}, {\bf B} and {\bf C} collect
the reference markers that are antecedent candidates
at the light only of the relevant binding constraints, which are
relative positioning filters in the process of anaphora resolution.%
\footnote{See \cite[Chap.2]{branco:diss99} for an overview
of filters and preferences for anaphora resolution proposed in the literature.
}
The elements in these list have to be submitted to the other constraints and preferences
of this process so that one of them ends up 
being chosen as the antecedent.

In particular, some of these markers 
may eventually turn up not being admissible antecedent candidates due to the violation of some
other constraints --- e.g. those requiring similarity of morphological features
or of semantic type --- that on a par with binding constraints
have to be complied with. For example, in this example {\it John described Mary to himself}, by the sole
constraining effect of Principle A, \mbox{[r-mark_{John}, r-mark_{Mary}]} 
is the list of antecedent candidates of
{\it himself}, which will be narrowed down to [r-mark_{John}] when all the other
filters for anaphora resolution have been taken into account, including
the one concerning similarity of morphological features,
as {\it Mary} and {\it him} do not have the same gender feature value.

In this particular case, separating these two type of filters --- similarity of
morphological features and binding constraints --- seems to be the correct
option, required by plural anaphors with so called split antecedents. In an example 
of this type, such as {\it John_{i} told Mary_{j} they_{i+j} would eventually get
married}, where {\it they} is resolved against {\it John} and {\it Mary}, the
morphological features of the anaphor are not identical to the morphological
features of each of its antecedents, though the relevant binding constraint
applies to each of them.%
\footnote{This was noted by \citep{higg:split83}. In this respect, this approach improves on the proposal in~\citep{polsag:hpsg94},
where the token-identity of indices --- internally structured in
terms of Person, Number and Gender features --- is meant to be
forced upon the anaphor and its antecedent in tandem with the relevant binding
constraint. 

For further reasons why token-identity between
the reference markers of the anaphor and the corresponding antecedent
is not a suitable option for every anaphoric dependency,
see the discussion below in Section \ref{discuss}
on the semantic representation of different modes of anaphora.}

When a plural anaphor takes more than one antecedent, as in the example above, 
its (plural) reference marker will end up being semantically related with a
plural reference marker resulting from some semantic combination
of the markers of its antecedents. Separating binding constraints from
other constraints on the relation between anaphors and their antecedents
are thus compatible with and justified by proposals for plural anaphora 
resolution that take into account split anaphora.%
\footnote{
That is the case e.g. of \citep{eschen:plural89}. According to this approach, the set of antecedent candidates of 
a plural anaphor which result from the verification of
binding constraints has to receive some expansion before subsequent 
filters and preferences apply in the anaphora resolution process. The reference
markers in that set, either singular or plural, will be 
previously combined into other plural reference markers: it is thus
from this set, closed under the semantic operation of 
pluralisation (e.g.\ i-sum a la~\citep{link:isums83}),
that the final antecedent will be chosen by the anaphor resolver.
}

\is{computational tractability}
\is{computational complexity}
\subsection{Computational tractability} 

It is also worth noting that the computational
tractability of the grammatical compliance with binding principles
is ensured given the polynomial complexity of the underlying operations
described above. 

Let {\it n} be
the number of words in an input sentence to be parsed, which for the
sake of the simplicity of the argument, and of the worst case scenario, 
it is assumed to be made only of nominal anaphors, that is every word in that sentence is a nominal anaphor. 
Assume also that the sets {\bf A}, {\bf Z} and {\bf U}, thus of length {\it n} at worst, 
are available at each node of the parsed tree via copying or via list appending 
(more details about these two operations in
the next sections), which is a process of constant time complexity. 

At worst, the operations involved at each one of the {\em n}
leaf nodes of the tree to obtain one of the sets {\bf A'}, {\bf Z'}, {\bf B} or {\bf C} are:
list copying and list appending operations, performed in constant time; extraction
of the predecessors of an element in a list, which is of linear complexity;
or at most one list complementation, which can be
done in time proportional to $n log(n)$. The procedure of verifying binding constraints in a sentence
of length {\it n} is thus of tractable complexity, namely $\mathcal{O}(n^2\log{}n)$ in the worst case.%
\footnote{
For a thorough discussion of alternative procedures for the compliance
with binding principles and their drawbacks, see \citep{branco:esslli2000}, very briefly summarised here:

The verification of binding constraints proposed in~\citep{chom:bind80,chom:lect81} requires extra-grammatical processing
steps of non tractable computational complexity \citep{correa:bind88, fong:index90}, which, moreover, 
are meant to deliver a forest of indexed
trees to anaphor resolvers. 

In Lexical Functional Grammar, the account of binding constraints requires special purpose 
extensions of the description
formalism \citep{dal:bind93}, which ensures only a partial handling
of these constraints.

For accounts of binding principles in  the family of Categorial Grammar frameworks, 
see \citep{szabol:89, hepple:90, morrill:2000}, 
and for a critical overview, see \citep{jaeger:2001}.
}

\section{Binding Constraints in the Grammar\label{spec1}}

In this section, the binding constraints receive
a principled integration into formal grammar. 
For the sake of brevity, 
we focus on the \ili{English} language. Given the discussion in the previous sections,
the parameterisation for other languages will follow from this example by means of seamless adaptation.

We show how the module of Binding Theory is specified 
with the description language of HPSG,
as an extension of the grammar fragment in the Annex of the foundational
HPSG book,\footnote{
\citep[Annex]{polsag:hpsg94}.}
following the feature geometry in Ivan Sag's proposed extension of this fragment to relative clauses,\footnote{
\citep{Sag97a}.} and adopting a semantic component 
for HPSG based on Underspecified Discourse Representation Theory (UDRT).\footnote{
\citep{frank:sem95}.}

As exemplified in (\ref{pronfeat}), this semantic component is encoded as the value of the 
feature {\sc cont(ent)}. This value, of sort {\em udrs}, has a structure permitting that the mapping 
into underspecified discourse representations be straightforward.\footnote{
\citep{reyle:udrt93}.}

The value of subfeature {\sc conds} is a set of labeled
semantic conditions. The hierarchical structure of
these conditions is expressed by means of a subordination relation
of the labels identifying each condition, a relation that
is encoded as the value of {\sc subord}. The attribute {\sc ls} defines
the distinguished labels, which indicate the upper ({\sc l-max}) and
lower ({\sc l-min}) bounds for a semantic condition within the overall semantic
representation to be constructed.

\textbf{{\sc anaph(ora)} subfeature of {\sc cont(ent)}} The integration of Binding Theory 
into formal grammar consists of a simple extension 
of this semantic component for the {\em udrs} of nominals,
enhancing it with the subfeature {\sc anaph(ora)}. This new feature keeps information 
about the anaphoric potential of the corresponding anaphor {\it w}.

Its subfeature {\sc antec(edents)} keeps record of how this potential 
is realised when the anaphor enters a grammatical construction: 
its value is the list with the antecedent candidates of {\it w} which 
comply with the relevant binding constraint for {\it w}. 

And its subfeature {\sc r(eference)-mark(er)} indicates the reference marker 
of {\it w}, which is contributed by its referential
force to the updating of the context.

\textbf{{\sc bind(ing)} subfeature of  {\sc loc(al)}} On a par with this extension 
of the {\sc loc} value, also the {\sc nonloc} value is extended 
with a new feature,
{\sc bind(ing)}, with subfeatures  {\sc list-a}, {\sc list-z}, and {\sc list-u}.
These lists provide a specification of the relevant context and
correspond to the lists {\bf A}, {\bf Z} and {\bf U} in the sections above. Subfeature {\sc list-lu}
is a fourth, auxiliary list encoding the contribution of the local 
context to the global, non local context, as explained in the next sections.%
\footnote{For the sake of readability, the working example in (\ref{pronfeat}) displays only 
the more relevant features for the point at stake. The {\sc nonloc} value has 
this detailed definition in \citep{polsag:hpsg94}:

\bigskip

\avmoptions{sorted}
\avmfont{\sc}
\avmvalfont{\it}
\avmsortfont{\it}
\begin{avm}
\[{nonloc} to-bind & nonloc1 \\
						     inherited & nonloc1 \]
\end{avm}

\bigskip

And these are the details of the extension we are using, where the information above
is coded now as a {\em udc} object, which 
keeps record of the relevant non local information for accounting
to {\em u(nbounded) d(ependency) c(onstructions)}:

\bigskip

\avmoptions{sorted}
\avmfont{\sc}
\avmvalfont{\it}
\avmsortfont{\it}
\begin{avm}
\[{nonloc}udc & \[{udc} to-bind & nonloc1 \\
						                  inherited & nonloc1 \]\\
										bind & \[{bind} list-a & list(refm) \\
																										list-z & list(refm) \\
																										list-u & list(refm) \\
																										list-lu & list(refm) \] \]
\end{avm}

\bigskip

Given this extension, HPSG principles constraining {\sc nonloc}
feature structure, or part of it, should be fine-tuned with adjusted
feature paths in order to correctly target the intended (sub)feature structures.
}

\subsection{Handling the anaphoric potential}

\textbf{Pronouns: lexical entry} Given this adjustment to the grammatical geometry, 
the lexical definition of a pronoun,
for instance, will include the following {\sc synsem} value:

\begin{exe}

\ex\label{pronfeat}
\avmoptions{active}
\avmfont{\sc}
\avmvalfont{\it}
\begin{avm}
[loc|cont & { [ ls & [l-max & @1\\
                       l-min & @1 ]\\
																	subord & \rm \{\} \\
																	conds & \{[label & @1\\
                        		dref & @2 ]\}\\
             				anaph & [r-mark & @2\\
                          antec & \it @5 principleB\ (@4,@3,@2)] ] }\\
  nonloc|bind & { [list-a & @3 \\
																										list-z & list\(refm\) \\
																										list-u & @4 \\
																										list-lu & <@2> \\] } ]
\end{avm}
\end{exe}

In this feature structure, the semantic condition in {\sc conds} associated
to the pronoun corresponds simply to the introduction of the discourse referent 
\raisebox{-.6ex}{\begin{avm}\@2\end{avm}}
as the value of {\sc dref}.

This semantic representation is expected
to be further specified as the lexical entry of the
pronoun gets into the larger representation of the
relevant utterance. In particular, the {\sc conds} value of the sentence
will be enhanced with a condition specifying the
relevant semantic relation between this reference marker
\raisebox{-.6ex}{\begin{avm}\@2\end{avm}}
and one of the reference markers in the value
\raisebox{-.6ex}{\begin{avm}\@5\end{avm}}
of {\sc antec}. The latter will be the antecedent against
which the pronoun will happen to be resolved, and the
condition where the two markers will be related represents the
relevant type of anaphora assigned to the anaphoric relation
between the anaphor and its antecedent.\footnote{
More details on the interface with anaphora resolvers and on the semantic types of anaphora in Section \ref{discuss}.}

The anaphoric binding constraint associated to pronouns, in turn,
is specified as the relational constraint \textit {principleB}/3
in the value of {\sc antec}.  
This is responsible for the realisation of
the anaphoric potential of the pronoun as it enters a grammatical
construction. When the arguments of this relational constraint are instantiated,
it returns list {\bf B} as the value of {\sc antec}. 

As discussed in Section \ref{semanticPatterns}, this relational constraint \textit {principleB}/3 
is defined to take 
all markers in the discourse context (in the first argument and given 
by the {\sc list-u} value), and remove from them both the local \mbox{o-commanders}  
of the pronoun (included in the second argument and made available by the {\sc list-a}
value)  and the marker corresponding to the pronoun (in the third  argument and
given by the {\sc dref} value).

Finally, the contribution of the reference marker of the pronoun
to the context is ensured via token-identity
between {\sc r-mark} and a {\sc list-lu} value.

The piling up of this reference
marker in the global {\sc list-u} value is determined
by a new HPSG principle specific to Binding Theory, to be detailed in the next Section \ref{contextRep}.

\textbf{Non pronouns and reflexives: lexical entries} The {\sc synsem} of other anaphors --- ruled by principles~A, C or Z --- are 
similar to the {\sc synsem} of pronouns above. The basic difference lies in the
relational constraints to be stated in the {\sc antec} value.

Such constraints  
---~{\it principleA}/2, {\it principleC}/3 and {\it principleZ}/2 --- encode the 
corresponding binding principles and return the realised
anaphoric potential of anaphors according to the surrounding context,
coded in their semantic representation under the form
of a list in the {\sc antec} value. Such 
lists --- {\bf A'}, {\bf C} or {\bf Z'}, respectively --- are obtained 
by these relational constraints along the lines
discussed in Section \ref{semanticPatterns}.

\textbf{Non lexical anaphoric expressions} Note that, for non-lexical anaphoric nominals in \ili{English}, namely those ruled
by Principle C, the binding constraint is stated in the lexical representation
of the determiners contributing to the anaphoric capacity of such NPs.
Also the reference marker corresponding to an NP of this kind is brought into 
its semantic representation from the {\sc r-mark} value
specified in the lexical entry of its determiner.

Accordingly, for the values of {\sc anaph} to be
visible in the signs of non lexical anaphors, 
Clause I of the Semantics Principle in UDRT\footnote{
\citep[p.12]{frank:sem95}.}
is extended with the
requirement that the {\sc anaph} value is token-identical,
respectively, with
the {\sc anaph} value of the specifier daughter, in an NP, and 
with the {\sc anaph} value of the nominal complement daughter, in 
a subcategorised PP.

\textbf{Exemption} Note also that for short-distance reflexives, exemption from the effect
of the corresponding Principle A occurs when
\raisebox{-0.085cm}{
\begin{avm}
{\it principleA}(\@3,\@2)
\end{avm}}
returns the empty list as the value of feature {\sc antec}:%
\footnote{
This account applies also to exempt occurrences of 
long-distance reflexives.}

\begin{exe}

\ex
\avmoptions{active}
\avmfont{\sc}
\avmvalfont{\it}
\begin{avm}
[loc|cont & { [ ls & [l-max & @1\\
                       l-min & @1 ]\\
																	subord & \rm \{\} \\
																	conds & \{[label & @1\\
                        		dref & @2 ]\}\\
             				anaph & [r-mark & @2\\
                          antec & \it @4 principleA\ (@3,@2)] ] }\\
  nonloc|bind & { [list-a & @3 \\
																										list-z & list\(refm\) \\
																										list-u & list\(refm\) \\
																										list-lu & <@2> \\] } ]
\end{avm}
\end{exe}

This happens 
if the reference marker of the reflexive \raisebox{-0.085cm}{
\begin{avm}
\@2
\end{avm}}
is the first element in the relevant
obliqueness hierarchy, i.e.\ it is the first element 
in the {\sc list-a} value in \raisebox{-0.085cm}{
\begin{avm}
\@3
\end{avm}},
thus \mbox{o-commanding} the other possible elements of this list and not being
\mbox{o-commanded} by any of them.

As discussed in Section \ref{Exemption}, given its essential anaphoricity,
a reflexive has nevertheless to be interpreted against some
antecedent. As in the exempt occurrences no antecedent candidate is 
identified by virtue of Principle A activation, the
anaphora resolver --- which will operate then on the empty {\sc antec} list%
\footnote{ 
More details of the
interface between grammar and reference processing systems in Section~\ref{discuss}.}
--- has thus to resort
to antecedent candidates outside the local domain of the
reflexive: this implies that it has to find
antecedent candidates for the reflexive which actually escape 
the constraining effect of Principle A. The anaphora resolver will then be
responsible for modelling the behaviour of reflexives 
in such exempt occurrences, in which case the
anaphoric capacity of these anaphors appears as being
exceptionally ruled by discourse-based factors.

\subsection{Handling the context representation}\label{contextRep}

\is{Binding Domains Principle}
Turning now to the representation of the context, this consists in the 
specification of the constraints on the values of the
attributes {\sc list-a}, {\sc list-z}, {\sc list-u}  and {\sc list-lu}. 
This is handled by adding
an HPSG principle to the grammar, termed 
the Binding Domains Principle (BDP). 
This principle has three
clauses constraining signs with respect to these four lists of reference
markers. A full understanding of their details, presented below, will be 
facilitated with the working example discussed in detail in the Appendix.

\textbf{Binding Domains Principle, Clause I} Clause I of BDP is responsible for ensuring
that the values of {\sc list-u} and \mbox{{\sc list-lu}} are appropriately
setup at the different places in a grammatical representation:

\begin{exe}
\ex\label{bdp}
\textbf{Binding Domains Principle}, Clause I
\begin{xlisti}
\ex The {\sc list-lu} value is identical to the 
concatenation of the {\sc list-lu} values of its daughters in every sign;
\ex  the {\sc list-lu} and {\sc list-u} values are 
token-identical in a sign of sort {\it discourse};
\ex
\begin{xlisti}
\ex  the {\sc list-u} value is token-identical to each 
{\sc list-u} value of its daughters in a non-NP sign;
\ex in an NP sign {\it k}:

\begin{itemize}
%\begin{xlisti}
\ex in Spec-daughter, the {\sc list-u} value is the result of removing the 
elements of the {\sc list-a} value of Head-daughter from the {\sc list-u} value of {\it k};
\ex in Head-daughter, the {\sc list-u} value is the result of removing 
the value of {\sc r-mark} of Spec-daughter from the {\sc list-u} value of {\it k}.
%\end{xlisti}
\end{itemize}

\end{xlisti}
\end{xlisti}
\end{exe}

\noindent By virtue of (i.), {\sc list-lu} collects up to the outmost sign in a grammatical 
representation --- which is of sort {\it discourse} --- the markers contributed to the context
by each NP. Given (ii.), this list with all the markers is passed to the
{\sc list-u} value at this outmost sign. And (iii.) ensures that this
list with the reference markers in the context is propagated to every NP. 

Subclause (iii.ii) prevents self-reference loops
due to anaphoric interpretation, avoiding what is known 
in the literature as the i-within-i effect --- recall that the {\sc r-mark} value of non lexical NPs 
is contributed by the 
lexical representation of their determiners, in Spec-daughter 
position, as noted above.
\is{i-within-i effect}

The HPSG top ontology is thus extended with the new subsort {\it discourse} 
for signs: $sign \equiv word \vee phrase \vee discourse$. This new type of 
linguistic object corresponds to sequences of sentential signs.
A new Schema 0 is also added to the
Immediate Dominance Principle, where the Head daughter is a 
phonologically null object of sort {\em context(ctx)}, and the Text daughter 
is a list of phrases. 

As the issue of discourse structure is out
of the scope of this paper, we adopted a very simple approach to the 
structure of discourses
which suffices for the present account of Binding Theory.
As discussed in the next \mbox{Section \ref{verif},} this object
of sort {\em ctx} helps representing the contribution of the non linguistic context to the 
interpretation of anaphors.

\textbf{Binding Domains Principle, Clause II} As to the other two Clauses of the Binding Domains Principle, 
Clause II and Clause III,
they constrain the lists {\sc list-a} and {\sc list-z}, respectively,
whose values keep a record of o-command relations.

BDP-Clause II is responsible for constraining {\sc list-a}:

\begin{exe}
\ex
\textbf{Binding Domains Principle}, Clause II
\begin{xlisti}
\ex	Head/Arguments: in a phrase, the {\sc list-a} value of its head, and of its nominal (or
nominal preceded by preposition) or trace Subject or Complement daughters are
token-identical;
\ex Head/Phrase:
\begin{xlisti}
\ex	in a non-nominal and non-prepositional sign, the {\sc list-a} values of a sign and its
head are token-identical;
\ex	in a prepositional phrase,
\begin{itemize}
\ex if it is a complement daughter, the {\sc list-a} values of the phrase and of its nominal
complement daughter are token-identical;
\ex otherwise, the {\sc list-a} values of the phrase and its head are token-identical;
\end{itemize}
\ex	in a nominal phrase,
\begin{itemize} 
\ex in a maximal projection, the {\sc list-a} value of the phrase and its Specifier daughter
are token-identical;
\ex in other projections, the {\sc list-a} values of the phrase and its head are
token-identical.
\end{itemize}
\end{xlisti}
\end{xlisti}
\end{exe}

This clause ensures that the {\sc list-a} value is shared between a head-daughter and its arguments, given (i.),
and also between the lexical heads and their successive projections, by virtue of (ii.).

\textbf{O-command} On a par with this Clause II, it is important to make sure
that at the lexical entry of any predicator {\it p}, {\sc list-a} includes 
the {\sc r-mark}
values of the  subcategorised arguments of {\it p} specified in its
{\sc arg-st} value. Moreover, the reference markers appear in the {\sc list-a} 
value under the same partial order as the order of the corresponding {\em synsem} 
in {\sc arg-st}. This is ensured by the following constraints on the lexical 
entries of predicators:

 \begin{samepage}
\begin{exe}
\ex\label{lexconst}
\end{exe}

\avmoptions{active}
\avmfont{\sc}
\avmvalfont{\it}
\begin{avm}
\hfill
\sort{{\it synsem}}{[loc|cont|arg-st <$\cdots$,
																																							 [loc|cont|anaph|r-mark & @i],$\cdots$>\\
	]}
\end{avm}
\begin{flushright}
$\longrightarrow$
\begin{avm}
\hfill
\sort{{\it synsem}}{[nonloc|bind|list-a <$\cdots$, @i,$\cdots$>]}
\end{avm}
\end{flushright}

\begin{avm}
\hfill
\sort{{\it synsem}}{[loc|cont|arg-st <$\cdots$, [loc|cont|anaph|r-mark & @k],\\
																																								$\cdots$,
																																						[loc|cont|anaph|r-mark @l],$\cdots$>
	]}
\end{avm}
\begin{flushright}
$\longrightarrow$
\begin{avm}
\hfill
\sort{{\it synsem}}{[nonloc|bind|list-a <$\cdots$, @k,$\cdots$, @l,$\cdots$>]}
\end{avm}
\\

\end{flushright}
\end{samepage}

In case a subcategorised argument is quantificational, 
it contributes also with its {\sc var} value to the make up of
{\sc list-a}:\footnote{
More details on this and on the
e-type anaphora vs. bound-variable anaphora distinction are discussed
in the next sections.}

\begin{exe}
\ex\label{lexconst2}
\end{exe}

\avmoptions{active}
\avmfont{\sc}
\avmvalfont{\it}
\begin{avm}
\hfill
\sort{{\it synsem}}{[loc|cont|arg-st <$\cdots$,
																																							 [loc|cont|anaph [r-mark & @r\\var &
@v]],$\cdots$>]}
\end{avm}
\begin{flushright}
$\longrightarrow$
\begin{avm}
\sort{{\it synsem}}{[nonloc|bind|list-a <$\cdots$, @v, @r,$\cdots$>]}
\end{avm}
\end{flushright}
\avmoptions{}

\pagebreak
\textbf{Binding Domains Principle, Clause III}  Finally, BDP-Clause III ensures that {\sc list-z} is properly
constrained:

\begin{samepage}
\begin{exe}
\ex
\textbf{Binding Domains Principle}, Clause III\\
For a sign F:
\begin{xlisti}
\ex	in a Text daughter, the {\sc list-z} and {\sc list-a} values are token-identical;
\ex	in a non-Text daughter,
\begin{xlisti} 
\ex in a sentential daughter, the {\sc list-z} value is the concatenation of the {\sc
list-z} value of F with the {\sc list-a} value; 
\ex in a Head daughter of a non-lexical nominal, the {\sc list-z} value is the concatenation of L
with the {\sc list-a} value, where L is the list which results from taking the list of
o-commanders of the {\sc r-mark} value, or instead of {\sc var} value when this exists,
of its Specifier sister from the {\sc list-z} value of F;
\ex in other, non-filler, daughters of F, the {\sc list-z} value is token-iden\-ti\-cal to
the {\sc list-z} value of F.
\end{xlisti}
\end{xlisti}

\end{exe}
\end{samepage}

By means of (i.), this Clause III ensures that, at the top
node of a grammatical representation, {\sc list-z} is set up as the
{\sc list-a} value of that sign. 

Moreover, given (ii.), it is ensured that {\sc list-z} is successively
incremented at suitable downstairs nodes --- those defining
successive locality domains for binding, as stated in (ii.i) and
(ii.ii) --- by appending, 
in each of these nodes, the {\sc list-a} value to the {\sc list-z} value
of the upstairs node.

\textbf{Locality} From this description of the Binding Domains Principle, it follows 
that the locus in grammar for the parameterisation 
of what counts as a local domain for a particular language is 
the specification of BDP--Clauses II and III for that language.

\is{anaphora resolution}
\is{reference processing}
\section{Interface with Reference Processing Systems \label{discuss}}

The appropriateness of the grammatical constraints on anaphoric binding presented above
extends to its suitable accounting of the division 
of labor between grammars and reference processing systems, and of the suitable interfacing between them.

\subsection{Anaphora Resolution \label{resolvers}}

While the grammatical anaphoric binding
constraints are specified and verified as part of the global set of grammatical
constraints, they provide also for a suitable hooking up of the grammar
with modules for anaphora resolution.

Feature {\sc antec} is the neat
interface point between them: its value with a list of antecedent
candidates that comply with Binding Theory requirements
is easily made accessible to anaphor resolvers. This list will be then handled by a
resolver where further non grammatical soft and hard constraints
on anaphora resolution will apply and will filter down that list
until the most likely candidate will be determined as the antecedent.

\subsection{Reference Processing\label{semanticTypes}}

The anaphoric binding constraints also provide a convenient interface for anaphoric links 
of different semantic types ---  exemplified below --- to be handled
and specified by reference processing systems:

\begin{exe}
\ex

\begin{xlist}

\ex {John$_{i}$ said that he$_{i}$ would leave soon.} (coreference)
\label{anTypes}

\ex {Kim$_{i}$ was introduced to Lee$_{j}$ and a few minutes later they$_{i+j}$ went off for dinner.} (split anaphora)
\label{anTypesb}

\ex {Mary could not  take [her car]$_{i}$ because [the tyre]$_{i}$ was flat.} (bridging anaphora)
\label{anTypesc}

\ex {[Fewer than twenty Parliament Members]$_{i}$ voted against the proposal because they$_{i}$ were afraid of riots in the streets.} (e-type anaphora)
\label{anTypesd}

\ex {[Every sailor in the Bounty]$_{i}$ had a tattoo with [his mother's]$_{i}$ name on the left shoulder.} (bound anaphora)
\label{anTypese}

\end{xlist}
\end{exe}

\is{coreference}
\is{plit antecedent}
\is{bridging anaphora}
\is{e-type anaphora}
\is{bound anaphora}
 Example (\ref{anTypes}) displays a coreference relation, where {\it he} has the same semantic value as its antecedent {\it John}. 
 
 A case of split antecedent can be found in (\ref{anTypesb}) as {\it they} has two syntactic antecedents and it refers to an entity comprising the two referents of the antecedents. 
 
 The referent of {\it the tyre} is part of the referent of its antecedent {\it his car} in (\ref{anTypesc}), 
 thus illustrating a case of so called bridging anaphora (also know as indirect or associative anaphora),
where an anaphor may refer to an entity that is e.g.\ an element or part 
of the denotation of the antecedent, or an entity that includes the denotation
of the antecedent, etc.% 
\footnote{
See~\citep{poesio:ana98} for an overview.
}

In (\ref{anTypesd}) {\it they} has 
a so called non-referential antecedent, {\it fewer than twenty Parliament Members}, 
from which a reference marker is inferred to serve as the semantic value of the plural pronoun: 
{\it they} refer to those Parliament Members, who are fewer than twenty in number, and who voted against the proposal.
 Example (\ref{anTypesd}) illustrates a case of e-type anaphora,\footnote{
\citep{evans:pron80}.} and this inference mechanism to obtain an antecedent marker
from a non referring nominal is described in Section~\ref{circAnaphPotential}. 

 Finally in (\ref{anTypese}), though one also finds a quantificational antecedent for the anaphoric expression, the relation of semantic dependency differs to the one in the previous example. The anaphoric expression {\it his mother} does nor refer to the mother of the sailors of the Bounty. It acts rather in the way of a bound variable of logical languages --- for each sailor $s$, {\it his mother} refers to the mother of $s$ --- thus exemplifying a case of so called bound anaphora.\footnote{
\citep{reinhart:bound83}.
}

Given that the semantic relation between antecedent marker and anaphor marker 
can be specified simply as another semantic condition added to the
{\sc conds} value, a DRT/HPSG representation for the resolved anaphoric
link under the relevant semantic type of anaphora is straightforward and the integration 
of the reference processing outcome into grammatical representation is seamlessly ensured.

For the sake of the illustration of this point, assume that a given reference 
marker {\bf x} turns out to be identified as the antecedent for the anaphoric
nominal Y, out of the set of antecedent candidates for Y in its {\sc antec}
value. This
antecedent {\bf x} can be related to the reference marker {\bf y} of anaphor Y by
means of an appropriate semantic condition in its {\sc conds} value. 
Such a condition will be responsible for modelling the specific mode of anaphora 
at stake.

For instance, coreference will require the
expected condition {\bf y}=_{coref}{\bf x}, as exemplified below 
with the {\sc cont} value of the pronoun in (\ref{pronfeat}) extended with a 
solution contributed by an anaphor resolver, where \raisebox{-0.7ex}{\em
\begin{avm}\@{7}\end{avm}} would be the marker picked up as the plausible 
antecedent.

\begin{exe}

\ex
\avmoptions{active}
\avmfont{\sc}
\avmvalfont{\it}
\begin{avm}
[ ls & [l-max & @1\\
                       l-min & @1 ]\\
																	subord & \rm \{@1=@6\} \\
																	conds & \{[label & @1\\
                        		dref & @2 ],
																											[label & @6\\
																												rel & $=$_{coref}\\
																												arg1 & @2\\
																												arg2 & @7]\}\\
             				anaph & [r-mark & @2\\
                          antec & @5<..., @7,...>] ] 

\end{avm}
\end{exe}

An instance of bridging anaphora, in turn,
may be modelled by {\it bridg}({\bf x}, {\bf y}), where {\it bridg} stands
for the relevant bridging function between {\bf y} and {\bf x}, and similarly 
for the other semantic anaphora types.

\subsection{Coreference Transitivity \label{transitivity}}

It is also noteworthy that the interfacing of grammar with reference processing systems
ensured by anaphoric binding constraints
provides a neat accommodation of coreference transitivity.

If as a result of the process of anaphora resolution, a given anaphor N and another anaphor B 
end up being both coreferent with a given antecedent A, then they end up being coreferent 
with each other. That is, in addition to having marker {\em r_{a}} as an admissible antecedent
in its set of candidate antecedents, that anaphor N has also to eventually have marker {\em r_{b}} 
included in that set.

This is ensured by including, in the {\sc conds} value in (\ref{pronfeat}), semantic conditions 
that follow as logical consequences from this overall coreference transitivity requirement that
is operative at the level of the reference processing system with which grammar is interfaced:
$\forall r_{a},r_{b} ((\raisebox{-.6ex}{\em \begin{avm}\@2\end{avm}}$=_{coref}$r_{b}
\wedge r_{a}$=_{coref}$r_{b})
\Rightarrow (\langle r_{a}\rangle \cup\raisebox{-.6ex}{\em
\begin{avm}\@5\end{avm}} = \raisebox{-.6ex}{\em
\begin{avm}\@5\end{avm}}))$.

An important side effect of this overall constraint is  that ``accidental"
violations of Principle B are prevented, as illustrated with the help of the following example.

 \begin{exe}
\ex [*]{The captain_{i/j} thinks he_{i} loves him_{j}.}
\end{exe}

Given that the Subject of the main clause, {\em the captain}, does not locally o-command any one of them,
either the pronoun {\em he} or the pronoun {\em him} can have the nominal phrase
{\em the captain} as antecedent, in compliance with Principle B. 
By transitivity of anaphoric coreference though, the reference marker 
of {\em he} is made to belong to the admissible set of antecedents of {\em him},
which violates Principle B. Hence, by the conjoined effect
of coreference transitivity and of Principle~B, that ``accidental" violation
of Principle B that would make {\em he} an (o-commanding)
antecedent of {\em him} in this example is (correctly) blocked.

By the same token, ``accidental" violations of Principle C 
with an analogous pattern as above, but for non pronouns, are prevented:

 \begin{exe}
\ex [*]{When John_{i/j} will conclude his therapy, [the boy_{i} will stop believing [that the patient_{j} is a Martian]].}
\label{transitivityC}
\end{exe}

Separately, {\em the boy} and the {\em the patient} can have {\em John} as antecedent, in accordance
to Principle C. 
But {\em the patient} --- because is o-commanded by {\em the boy} --- cannot have {\em the boy} as antecedent, 
which, also here, is (correctly) ensured by a conjoined 
effect of the coreference transitivity requirement and the relevant Principle~C.

Accordingly, when the semantic type of anaphora is not one of coreference,
no coreference transitivity holds, and there happens no ``accidental" violation of Principle~C.
This is illustrated in the following example with bridging anaphora instead,
where two non pronouns, though occurring in the same clause, like in (\ref{transitivityC}) , can be (correctly) resolved against 
the same antecedent --- in contrast with that example (\ref{transitivityC}) above, where such possibility is blocked.

\begin{exe}
\ex
\gll Quando [o robot]_{i} concluiu a tarefa, o operador viu que [a roda]_{i} estava a esmagar [o cabo de alimenta\c{c}\~ao]_{i}. (\ili{Portuguese})\\ 
when the robot concluded the task, the operator saw that the wheel was to crush the cord of power\\
\trans 'When [the robot]_{i} concluded the task, the operator saw that [his_{i} wheel] was crushing [his_{i} power cord].'
\end{exe}

Another range of examples where the semantic type of anaphora is not one of coreference --- also with
no coreference transitivity holding --- and thus also where (correctly) there happens no``accidental" violation
of the respective binding principle can be found for reflexives, as illustrated in the following example.

 \begin{exe}
\ex {The captain_{i} thinks he_{i/j} loves himself_{*i/j}.}
 \label{accidentalReflexives}
\end{exe}

The reflexive {\em himself} can have {\em he} as antecedent, because the later locally
o-commands it, but cannot have {\em the captain} as antecedent because the later
does not locally o-command it. But while the semantic anaphoric relation between {\em the captain} 
and {\em he} is one of coreference, the semantic anaphoric relation between {\em he}
and {\em himself} is not, being rather one of bound anaphora.%
\footnote{
Confluent evidence that reflexives entertain a bound anaphora relation with their antecedents
was also observed when their inability to enter split anaphora relations in non exempt positions
was noted in Section \ref{Exemption}.
}
Hence, 
the coreference transitivity requirement does not apply and the referent
of {\em the captain} does not land into the set of possible antecedents
of the reflexive, thus not inducing an ``accidental" violation of Principle A. Example  (\ref{accidentalReflexives})
can thus felicitously be interpreted as the captain thinking that the agent of loving him is himself,
resulting from {\em himself} having {\em him} as antecedent and {\em him} having {\em the captain} as antecedent.

\section{Binding Constraints for Antecedents \label{reverse}}

The Binding Theory presented in this paper is also 
serendipitous in terms of improving the accuracy of empirical predictions offered 
by a formal grammar with respect to anaphoric binding restrictions that are outside the
realm of the binding principles in (\ref{PrincipleA})-(\ref{PrincipleC}).

\is{e-type anaphora}
\is{bound anaphora}
Note first that a reference marker introduced by a non quantificational NP can be 
the antecedent either of an anaphor that it o-commands, as in  (\ref{nonquantOcom}), or of an anaphor that
it does not o-command, as in (\ref{nonquantNotOcom}):

\begin{exe}
\ex
\begin{xlist}
 \label{nonquantOcom}
\ex[]{$[$The captain who knows this sailor]_{i} thinks Mary loves him_{i}.}
 \label{nonquantOcom}
\ex[]{$[$The captain who knows [this sailor]_{i}] thinks Mary loves him_{i}.}
 \label{nonquantNotOcom}
\end{xlist}
\end{exe}

%The two markers introduced by a quantificational NP, in turn, present a different behaviour. 

Differently from a non quantificational NP, which contributes
one reference marker to the representation of the context, a quantificational
NP contributes two markers that exhibit symmetric features
with respect to each other in several respects. 
%This has important consequences in terms of the specific binding 
%capacity of such markers (and of the respective nominal expressions introducing them)
%as antecedents of anaphors. 
The fact that
one of them can serve as an antecedent in e-type anaphora, while
the other can serve as an antecedent in bound-variable anaphora is certainly
one of such symmetries.\footnote
{Extensive discussion  of this difference is presented in the Appendix.}
 But there are more.

Let us take a quantificational NP, introduced for instance by the quantifier {\em every}, 
acting as an antecedent. This imposes
different Number requirements on its anaphors depending on the
type of anaphora relation at stake --- e-type or bound-variable anaphora --- so that the
underlying occurrence of each one of the corresponding
two markers can be tracked down.

%For ease of reference, let us
%refer to the marker in the {\sc r-mark} value, introduced by 
%\mbox{\textSigma-abstraction}, 
%as the e-marker, and to the marker in the the {\sc var} value,
%introduced by the restrictor argument of the determiner, as the v-marker.

For ease of reference, let us term the marker ensuring e-type anaphora as the e-marker, 
and the marker ensuring bound anaphora as the v-marker.\footnote{
In the formalisation presented in the Appendix, an e-marker is the marker in the {\sc r-mark} value, 
introduced by \mbox{\textSigma-abstraction}, 
and a v-marker is the marker in the the {\sc var} value,
introduced by the restrictor argument of the determiner.}

The contrast below illustrates that, in an e-type anaphoric link, the e-marker stands for a plurality:

\begin{exe}
\ex[]{Every sailor_{i} has many girlfriends. They_{i}/He_{*i} travel(s) a lot.}
\end{exe}

And the next contrast illustrates that, in a bound-variable anaphoric link, the v-marker is singular:

\begin{exe}
\ex[]{Every sailor_{i} shaves themselves_{*i}/himself_{i}.}
\end{exe}

The following contrasts can now be considered. An e-marker can be the
antecedent of anaphors that it does not o-command, in (\ref{etypeb}), but
cannot be the antecedent of anaphors that it o-commands, in (\ref{etypea}):

\begin{exe}
\ex\label{etype}
\begin{xlist}
\ex[*]{$[$Every captain who knows this sailor]_{i} thinks Mary loves them_{i}.}
\label{etypea}
\ex[]{$[$The captain who knows [every sailor]_{i}] thinks Mary loves them_{i}.}
\label{etypeb}
\end{xlist}
\end{exe}

This contrast is symmetric to the contrast for the other reference marker: a v-marker 
can be the antecedent of anaphors that it o-commands, in (\ref{weakcrossovera}), but
cannot be the antecedent of anaphors that it 
does not o-command, in (\ref{weakcrossoverb}):

\begin{exe}
\ex\label{weakcrossover}
\begin{xlist}
\ex[]{$[$Every captain who knows this sailor]_{i} thinks Mary loves him_{i}.}
\label{weakcrossovera}
\ex[*]{$[$The captain who knows [every sailor]_{i}] thinks Mary loves him_{i}.}
\label{weakcrossoverb}
\end{xlist}
\end{exe}

As these contrasts are empirically observed as patterns holding for quantificational NPs
in general (not only for those introduced by {\em every}), 
constraints emerge on which anaphors different markers can be the antecedents 
of, in case such markers are contributed by quantificational NPs.

E-markers and v-markers of a given quantificational NP induce a partition of the
space of their possible anaphors when that NP is acting as an
antecedent:  a~\mbox{v-marker}
is an antecedent for anaphors in the set of its o-commanded anaphors, while  
an e-marker is an antecedent for anaphors in the complement of such set, i.e.\ in 
the set of its non o-commanded anaphors.

This implies that on a par with the grammatical constraints on {\em the relative
positioning of antecedents with respect to anaphors} in (\ref{PrincipleA})-(\ref{PrincipleC}), 
there are also grammatical constraints on {\em the relative positioning 
of anaphors with respect to their antecedents} when
the corresponding markers are introduced by quantificational NPs. 
Building on the same auxiliary notions, these ``reverse" binding
constraints receive the following definition as R-Principles E and V:

\is{R-Principle V}
\is{R-Principle E}

\begin{exe}
\ex\label{reverseBindingPrinciples}
{\textbf{R-Principle E:}} An antecedent cannot o-bind its anaphor (in \mbox{e-type} anaphora).

\sn
{$[$Every captain who knows [every sailor]_{i}]_{j} thinks Mary loves them_{i/*j}.}
\end{exe}

\begin{exe}
\sn
{\textbf{R-Principle V:}} An antecedent must o-bind its anaphor (in bound-anaphora).

\sn
{$[$Every captain who knows [every sailor]_{i}]_{j} thinks Mary loves him_{*i/j}.}
\end{exe}

It is worth noting that these principles account also for what has been observed in the literature 
as the weak crossover effect.%
\footnote{See \citep[Sec.2.1]{jacobson:paycheck2000} for an extensive overview of 
accounts of weak crossover. For an account of strong crossover in HPSG see
\citep[p.279]{polsag:hpsg94}. }
In the example below, displaying a case
of weak crossover, the anaphoric link is ruled out
by \mbox{R-Principle V} since the quantificational NP {\em every sailor} does not
o-command the pronoun {\em him}, which is singular and could thus enter
only into a bound-anaphora relation. 

\begin{exe}
\ex[*]{$[$The captain who knows him_{i}] thinks Mary loves every sailor_{i}.}
\end{exe}

Weak crossover constructions
appear thus as a sub-case of the class of constructions ruled out
by the binding constraints for antecedents.%
\footnote{
To the best of our knowledge, the integration of the reverse anaphoric constraints 
E and V in (\ref{reverseBindingPrinciples}) into HPSG ---
like what is obtained in Section \ref{spec1} above for Principles A-Z in (\ref{PrincipleA})-(\ref{PrincipleC}) ---
was not worked out yet in the literature.

Besides an explicit formal specification of (\ref{reverseBindingPrinciples}) in terms of HPSG, 
there are also empirical aspects
that ask to be worked out in future work. 
For weak crossover, for instance, it is interesting to note Jacobson's remarks: ``... it is well known
that weak crossover (WCO) is indeed weak, and that the effect can be
ameliorated in a variety of configurations. To list a few relevant
observations: WCO violations are much milder if the offending
pronoun is within a sentence rather than in an NP; the more deeply
one embeds the offending pronoun the milder the WCO effect;
WCO effects are ameliorated or even absent in generic sentences;
they are milder in relative clauses than in questions [...] For example,
the possibility of binding in {\em Every man's_{i} mother loves him_{i}}
remains to be accounted for." \citep[p.120]{jacobson:paycheck2000}.
}

\section{Outlook \label{outlook}}

With the material presented in the sections above, it emerges that the grammar 
of anaphoric binding constraints builds on the following key ingredients:

\begin{itemize}

\item Interpretation: binding constraints are grammatical constraints on interpretation 
contributing to the contextually determined
semantic value of anaphors --- rather than syntactic wellformedness constraints.

\item Lexicalisation: binding constraints are properties of 
anaphors determining how their semantic value 
can be composed or co-specified, under a non-local syntactic geometry, with the
semantic value of other expressions --- rather than properties of grammatical
representations of sentences as such: accordingly, the
proper place of these constraints in grammar is at the lexical description of 
the relevant anaphoric units (e.g. the English pronoun {\em him}, or the Portuguese multiword long distance reflexive {\em ele pr\'{o}prio}) or the anaphora inducing items (e.g. the English definite article {\em the} that introduces non-pronouns).

\item Underspecification: binding constraints delimit how the
anaphoric potential of anaphors can be realised when they
enter a grammatical construction --- rather than determining the eventual antecedent: 
on the one hand, this realisation of anaphoric potential is not a final solution in terms of circumscribing
the elected antecedent, but a space of grammatically admissible solutions;
on the other hand, this realisation of anaphoric potential has to be decided, locally,
in terms of non-local information: accordingly, an underspecification-based
strategy is required to pack ambiguity and non-locality.

\item Articulation: binding constraints are grammatical constraints --- rather than 
anaphora resolvers: accordingly, grammars, where grammatical ana\-phoric
constraints reside, and reference processing systems, where further
constraints on the resolution of anaphora reside,
are autonomous with respect to each other, and their specific contribution gains
from them being interfaced, rather than being mixed up.
\end{itemize}

Binding principles capture the relative positioning of anaphors and their 
admissible antecedents in grammatical representations. As noted at the introduction of the present paper, together with their auxiliary notions, they have been considered one of the most outstanding modules 
of grammatical knowledge.

From an empirical perspective, these constraints stem from quite cogent generalisations and exhibit a universal
character, given the hypothesis of their parameterised validity across anaphoric expressions and natural languages.

From a conceptual point of view, in turn, the relations among binding
constraints involve non-trivial cross symmetry that lends them a modular nature
and provides further strength to the plausibility of their universal character.

To conclude the overview presented in this paper, the remainder two subsections below present intriguing and promising research questions, respectively for symbolic and neural approaches.

\subsection{Symbolic}

\textbf{Symmetries}
The recurrent complementary distribution of the admissible antecedents
of a pronoun and of a short-distance reflexive in the same, non exempt
syntactic position, in different languages from different language families,
has perhaps been the most emblematic symmetry. 

For the sake of convenience,
the examples in (\ref{PrincipleA})-(\ref{PrincipleC}) are copied
to (\ref{exA})-(\ref{exC}) below. The pair (\ref{exA}) vs. (\ref{exB}), with the anaphoric expressions
in the same syntactic position of the same syntactic construction, illustrates
the symmetry just mentioned, between reflexives and pronouns, suggestively grasped by comparing the starred 
and non starred indexes.

\pagebreak
\begin{exe}
\ex
\label{exA}
{...{\em X}$_{x}$...[Lee$_{i}$'s friend]$_{j}$ thinks
[[Max$_{k}$'s brother]$_{l}$ likes %\linebreak
himself$_{*x/*i/*j/*k/l}$].}

\ex
\label{exZ}
\gll ...{\em X}$_{x}$...[O amigo do Lee$_{i}$]$_{j}$ acha [que [o
irm\~{a}o do Max$_{k}$]$_{l}$ gosta dele pr\'{o}prio$_{*x/*i/j/*k/l}$]. (\ili{Portuguese})\\ \mbox{ }\mbox{ }\mbox{ }\mbox{ }\mbox{ }\mbox{ }\mbox{ }\mbox{ }\mbox{ }\mbox{ }\mbox{ }\mbox{ }the friend of.the Lee thinks \mbox{ }that \mbox{ }the brother of.the Max likes of.him self\\
\trans '...{\em X}$_{x}$...[Lee$_{i}$'s friend]$_{j}$ thinks [[Max$_{k}$'s brother]$_{l}$
likes him$_{*x/*i/j/*k}$ / \linebreak
himself$_{l}$].'

\ex
\label{exB}
{...{\em X}$_{x}$...[Lee$_{i}$'s friend]$_{j}$ thinks [[Max$_{k}$'s brother]$_{l}$
likes %\linebreak
 him$_{x/i/j/k/*l}$].}

\ex
\label{exC}
{...{\em X}$_{x}$...[Lee$_{i}$'s friend]$_{j}$ thinks [[Max$_{k}$'s brother]$_{l}$
likes %\linebreak 
the boy$_{x/i/*j/k/*l}$].}
\end{exe}
%\vspace{4 mm}

But given also the complementary 
distribution of the admissible antecedents of a long-distance
reflexive and of a non pronoun in the same, non exempt syntactic position,
a similar symmetry is also found between these two other types of anaphors.
This is illustrated by the complementarity of the indexes in (\ref{exZ}) vs. (\ref{exC}).

Another double ``symmetry'' worth noting is the one
between short- and long-distance reflexives, on the one hand,
and non pronouns and pronouns on the other hand.

Both sorts
of reflexives present the same binding regime but over
o-command orders whose length is possibly different: the set
of admissible antecedents of a short-distance reflexive is a subset
of the set of admissible antecedents of a long-distance reflexive
in the same, non exempt syntactic position. For a given non-exempt position,
the admissible antecedents of a short-distance reflexive are the antecedents 
that are in the set of admissible antecedents of a long-distance reflexive
in that same position and that are local, i.e. are in the local domain. The felicitous (non starred) indexes 
in (\ref{exA}) are a subset of the felicitous indexes in (\ref{exZ}), which illustrates this symmetry.

A ``symmetry" similar to
this one is displayed by non pronouns and pronouns with respect
to a given syntactic position: the set of admissible antecedents of
a non pronoun is a subset of the set of admissible antecedents of
a pronoun. For a given position,
the admissible antecedents of a non pronoun are the antecedents 
that are in the set of admissible antecedents of a pronoun
in that same position and that are not o-commanding the pronoun (or non pronoun).
The felicitous (non starred) indexes 
in (\ref{exC}) are a subset of the felicitous indexes in (\ref{exB}).

\textbf{Quantificational Strength}
When these symmetries are further explored, 
the intriguing observation that emerges 
with respect to the empirical generalisations in
(\ref{PrincipleA})-(\ref{PrincipleC}) is that when stripped away from
their procedural phrasing and non-exemption safeguards,
they instantiate a square of logical oppositions:

\is{square of opposition}
\begin{exe}
\ex
\label{bindingSquareOpposition}
\end{exe}
\vspace{-7mm}
\centerline{\includegraphics[width=18pc]{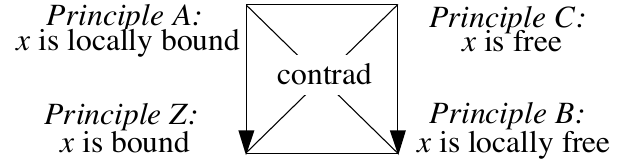}}

\is{square of opposition}
\is{square of duality}

Like in the Aristotelian square of opposition, depicted in (\ref{patternSquareOpposition}), there are two pairs of {\em contradictory} constraints, which are formed
by the two diagonals, (Principles A, B) and (C, Z). One pair of {\em contrary}
constraints (they can be both false but cannot be both true) is given
by the upper horizontal edge (A, C).  One pair of {\em compatible}
constraints (they can be both true but cannot be both false) is given
by the lower horizontal edge (Z, B). Finally two pairs of
{\em subcontrary} constraints (the first coordinate implies the second,
but not vice-versa) are obtained by the vertical edges, (A, Z) and (C,
B).

\begin{exe}
\ex
\label{patternSquareOpposition}
\end{exe}
\vspace{-10mm}
\centerline{\includegraphics[width=13pc]{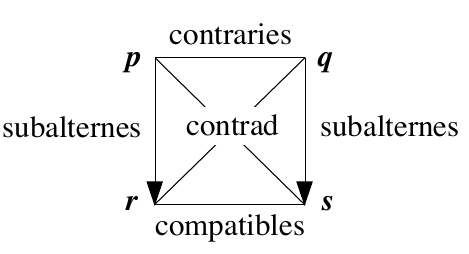}}

The empirical
emergence of a square of oppositions for the semantic values of natural
language expressions naturally raises the question about the possible
existence of an associated square of duality --- and importantly,
about the quantificational nature of these expressions.

\is{square of duality}
\is{quantification}
\is{quantifier}
\begin{exe}
\ex
\label{patternSquareDuality}
\end{exe}
\vspace{-7mm}
\centerline{\includegraphics[width=12pc]{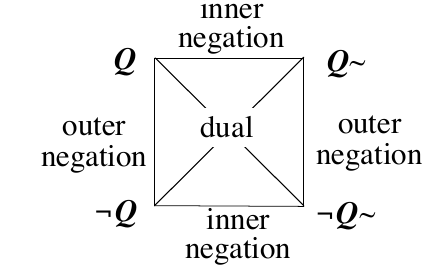}}

It is of note that 
the classical square of oppositions in (\ref{patternSquareOpposition})
is different and logically independent from
the square of duality in (\ref{patternSquareDuality}) --- with the semantic values of the English 
expressions {\em every N}, {\em no N},
{\em some N} and {\em not every N}, or their translational equivalents in
other natural languages, providing the classical example of an instantiation
of the latter:

\begin{exe}
\ex
\end{exe}
\vspace{-7mm}
\centerline{\hspace{0 mm}\includegraphics[width=14pc]{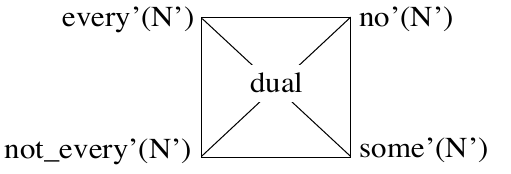}}

The difference lies in the fact that inner negation, outer
negation and duality (concomitant inner and outer negation) are third order concepts, while compatibility, contrariness and
implication are second order concepts. As a consequence, it is possible
to find instantiations of the square of oppositions without a
corresponding square of duality, and vice-versa.%
\footnote{
Vd.~\citep{Lobner1987} for examples and discussion.
}

Logical duality has been a key issue in the study of natural language 
and, in particular, in the study of quantification as this happens to
be expressed in natural language.  It is a pattern noticed in the
semantics  of many linguistic expressions and phenomena, ranging from
the realm  of determiners to the realm of temporality and modality,
including topics  such as the semantics of the adverbials {\em
still}/{\em already} or of the conjunctions  {\em because}/{\em
although}, etc.%
\footnote{
\citep{Lobner1987, Lobner1989, Lobner1999, terMeulen1988, Konig1991, Smessaert1997}.
}

Under this pattern, one recurrently finds groups of syntactically related 
expressions whose formal semantics can be rendered as one of the operators
arranged in a square of duality. Such a square is made of operators that 
are interdefinable by means of the relations of outer negation, inner
negation, or duality.
Accordingly, 
the emergence of a notoriously non trivial square of logical duality
between the semantic values of natural language expressions has been taken 
as a major empirical touchstone to ascertain their quantificational nature.%
\footnote{
Vd. \citep{Lobner1987,vanBenthem1991}. While noting that the ubiquity of the square of duality 
may be the sign of a semantic invariant possibly rooted in some cognitive universal,  
\citep[p.23]{vanBenthem1991} underlined its heuristic value for research
on quantification inasmuch as ``it suggests a systematic point of view from
which to search for comparative facts".
}

By exploring these hints, and motivated by the intriguing square of opposition in (\ref{bindingSquareOpposition}), 
the empirical generalisations captured in the binding principles were shown to be the effect 
of four quantifiers that instantiate a square of duality like (\ref{patternSquareDuality}).%
\footnote{
%\citep{branco:1998,branco:2001,branco:2005,branco:2006}.
\citep{branco:2006,branco:2005,branco:2001,branco:1998}.
}

\is{Principle A}
\is{Principle Z}
\is{Principle B}
\is{Principle C}
\is{reflexives}
\is{short-distance reflexives}
\is{long-distance reflexives}
\is{pronouns}
\is{non-pronouns}
\is{phase quantification}

For instance, Principle A is shown to capture the constraining effects of the existential quantifier 
that is part of the semantic value of short-distance reflexives. Like the existential
quantifier expressed by other expressions, such as the adverbial {\em already},%
\footnote{
\citep{Lobner1987}.
}
this a phase quantifier. What is specific here is that the quantification is over
a partial order of reference markers,  the two relevant semi-phases over this order
include the local o-commanders and the other reference markers that are not
local o-commanders, respectively for the positive and the negative semi-phases,
and the so-called parameter point in phase quantification is the reference
marker of the eventual antecedent for the anaphoric nominal at stake.

Accordingly, the other
three quantifiers --- corresponding to the other three binding Principles B, C and Z --- 
are defined by means of this existential one being under external negation (quantifier expressed by pronouns), 
internal negation (by non pronouns) or both external and internal negation (by long-distance reflexives).

\is{referential nominals}
\is{quantificational nominals}
\is{dual nominals}
\is{e-type anaphora}

\textbf{Doubly Dual Nominals}
While these findings deepen the rooting of binding constraints into the semantics of anaphoric nominals,%
\footnote{
Their fully-fledged discussion and justification are outside the scope of the present paper. A thorough presentation can be found in \citep{branco:2005}.} more importantly, they also point towards promising research directions
with the potential to advance our understanding of the grammar of anaphoric binding, in particular, 
and more widely, to further our insights into the semantics of nominals, in general.

A shared wisdom is that nominals convey either quantificational or referential
force. 

The findings introduced above imply that nominals with ``primary" referential force 
(e.g. {\em John}, {\em the book}, {\em he},...) have also a certain ``secondary" quantificational force: 
they express quantificational requirements --- over reference markers, i.e. entities that live in linguistic 
representations ---, but do not directly  quantify  over  extra-linguistic  
entities,  like  the  other  ``primarily" quantificational nominals 
(e.g. {\em every man}, {\em most students},...) do.

This  duality  of  semantic  behaviour,  however,  turns  out  not  to  be  that  much
surprising  if  one  takes into account  a  symmetric  duality  with  regards  to  ``primarily" quantificational
nominals, which is apparent when they are able to act as antecedents in e-type anaphora. 
Nominals  with  ``primary"  quantificational 
force have also a certain ``secondary" referential force: they
have enough referential strength to evoke and introduce reference markers in the
linguistic representation that can be picked as antecedents by anaphors --- and
thus support the referential force of the latter~---, but they cannot be used to directly refer 
to extra-linguistic entities, like the other ``primarily" referential terms do.

As a result, the duality quantificational vs. referential nominals appears thus as less strict and more articulated 
than it has been assumed. Possibly
taking indefinite descriptions aside, every nominal makes a contribution in both semantic  dimensions  
of  quantification  and  reference  but  with  respect  to  different
universes. Primarily referential nominals have a dual semantic nature --- they are
primarily referential (to extra-linguistic entities) and secondarily quantificational 
(over linguistic entities) ---, which is symmetric of
the  dual semantic  nature  of  primarily  quantificational  ones --- these  are  primarily
quantificational (over extra-linguistic entities) and secondarily referential (to linguistic
entities).

\subsection{Neural}

\textbf{Natural Language Processing Task} Some natural language processing tasks, e.g. question answering, appear as end to end procedures serving some useful, self-contained application. Some other tasks, in turn, e.g. part-of-speech tagging, appear more as instrumental procedures to support those downstream, self-contained applications. To help assess research progress in neural natural language processing, sets of processing tasks, of both kinds, have been bundled together, e.g. in the GLUE benchmark.%
\footnote{\citep{wang-etal-2018-glue}.}

As one such instrumental natural language processing task, possibly contributing or being embedded into downstream applications, anaphora resolution, including coreference resolution, is a procedure by means of which anaphors are paired with their antecedents. It has been addressed with neural approaches%
\footnote{\citep{lee-etal-2017-end,xu-choi-2020-revealing}.}
and has been integrated into natural language processing benchmarks.

While related to anaphora resolution, and eventually instrumental to it, determining the set of grammatically admissible antecedents for a given anaphor is a procedure that, as such, has not been addressed yet with neural approaches, to the best of our knowledge.
Like many other instrumental tasks, this is a challenge that can contribute to make empirically evident and to appreciate the strength of different neural approaches in handling natural language processing. Grammatical anaphoric binding is thus an intriguing research question open to be addressed with neural approaches, and also with a good potential to provide a research challenge that may pave the way for neuro-symbolic solutions to emerge.

\textbf{Probing for Linguistic Plausibility} While providing outstanding performance scores in many natural processing tasks, neural models have been challenged, like in other applications areas, due to its opacity and lack of interpretability, specially when compared to symbolic methods.

As a way to respond to this type of challenge, neural models have been submitted to ingenious probing procedures
aimed at assessing them with respect to the linguistic knowledge they may eventually have specifically encoded while having been trained for generic or high level natural language processing tasks, like for instance language modelling, machine translation, etc.%
\footnote{\citep{conneau-etal-2018-cram,tenney-etal-2019-bert,miaschi-dellorletta-2020-contextual}, \textit{i.a.}}

This endeavour of unveiling the possible linguistic knowledge represented in neural models will certainly benefit from integrating the task of grammatical anaphoric binding in this kind of toolboxes that may be used for linguistic probing and interpretability. 

\textbf{Inductive Bias for Natural Language} An increasingly important research question in neural natural language processing is to design models that possibly have an appropriate inductive bias such that their internal linguistic representations and capabilities resemble as much as possible the ones of human language learners after being exposed with as little volume of raw training data as the ones humans learners are exposed to.%
\footnote{\citep{mccoy-etal-2020-syntax}.}

A most outstanding feature of natural language is the possibility of there being so called long distance relations, that is relations between expressions among which a string of other expressions of arbitrary length may intervene. This builds on another feature that has been widely recognized as underlying natural language, namely the hierarchical nature of its complex expressions.%
\footnote{\citep{chomsky:1965}}

As amply documented in the overview above, grammatical anaphoric binding relations, among anaphors and antecedents, are grammar regulated connections that are long distance relations par excellence. Hence, anaphoric binding is essential, and of utmost importance, for the endeavour of designing neural models with appropriate inductive bias for natural language.

\appendix
\section*{Appendix \label{verif}}
%\addcontentsline{toc}{section}{Appendix}
%\renewcommand{\thesubsection}{\Alph{section}}
\setcounter{section}{1}

In order to illustrate the combined effect of the binding constraints specified in Section \ref{spec1}, 
as well as the outcome obtained from a grammar that integrates Binding Theory,
we work through the example below and the corresponding grammatical
representation in Figure 1. %
%\footnote{
%In order to check its consistency with the rest of the grammar
%and its practical viability, the formal specification of binding theory 
%presented in the previous Section was coded into
%the computational implementation of an HPSG fragment grammar 
%similar to the one presented in the Appendix of~\citep{polsag:hpsg94}.
%For the sake of an agile implementation, we resorted to Prolog
%and to its extension ProFIT~\citep{erbach:profit95}
%as the implementation formalism.
%The relational constraints expressing binding principles were implemented by
%means of Prolog predicates associated with the lexical clauses of 
%anaphoric expressions. These relational constraints were
%defined in terms of simple auxiliary predicates ensuring the basic
%operations of list appending, list difference, etc.
%
%Some of the auxiliary predicates used for the implementation
%of binding constraints have arguments,
%e.g {\sc list-u}, whose value is obtained when the whole
%relevant grammatical representation is built up. As expected, this is a consequence
%of accumulating and packing non-local information in such lists. As in Johnson's
%approach \citep(johnson:bind95), this requires that some delaying device be used
%in the activation of those predicates, which in this implemented
%grammar was done by resorting to the Prolog built-in predicate \texttt{freeze/2}.
%}

\begin{exe}
\ex{\label{ex:topic}}{Every student said [he likes himself].}
\end{exe}

This is a
multi-clausal sentence with two anaphoric nominals in the embedded
clause, a pronoun ({\em he}) and a short-distance reflexive ({\em himself}), and with a quantificational
NP ({\em every student}) in the upper clause. In this sentence,
the reflexive has the pronoun as the only admissible antecedent,
and the pronoun, in turn, can either have the quantificational
NP as antecedent or be resolved against an antecedent not 
introduced in the sentence.

Figure 1 presents an abridged version of the grammatical 
representation produced by the grammar for a discourse
that contains only this sentence.
The feature structures below the constituency tree correspond to partial grammatical 
representations of the leave constituents, while the ones above the 
tree correspond to partial representations of some of its non terminal nodes.

\begin{figure}
\begin{center}
\includegraphics[width=12cm]{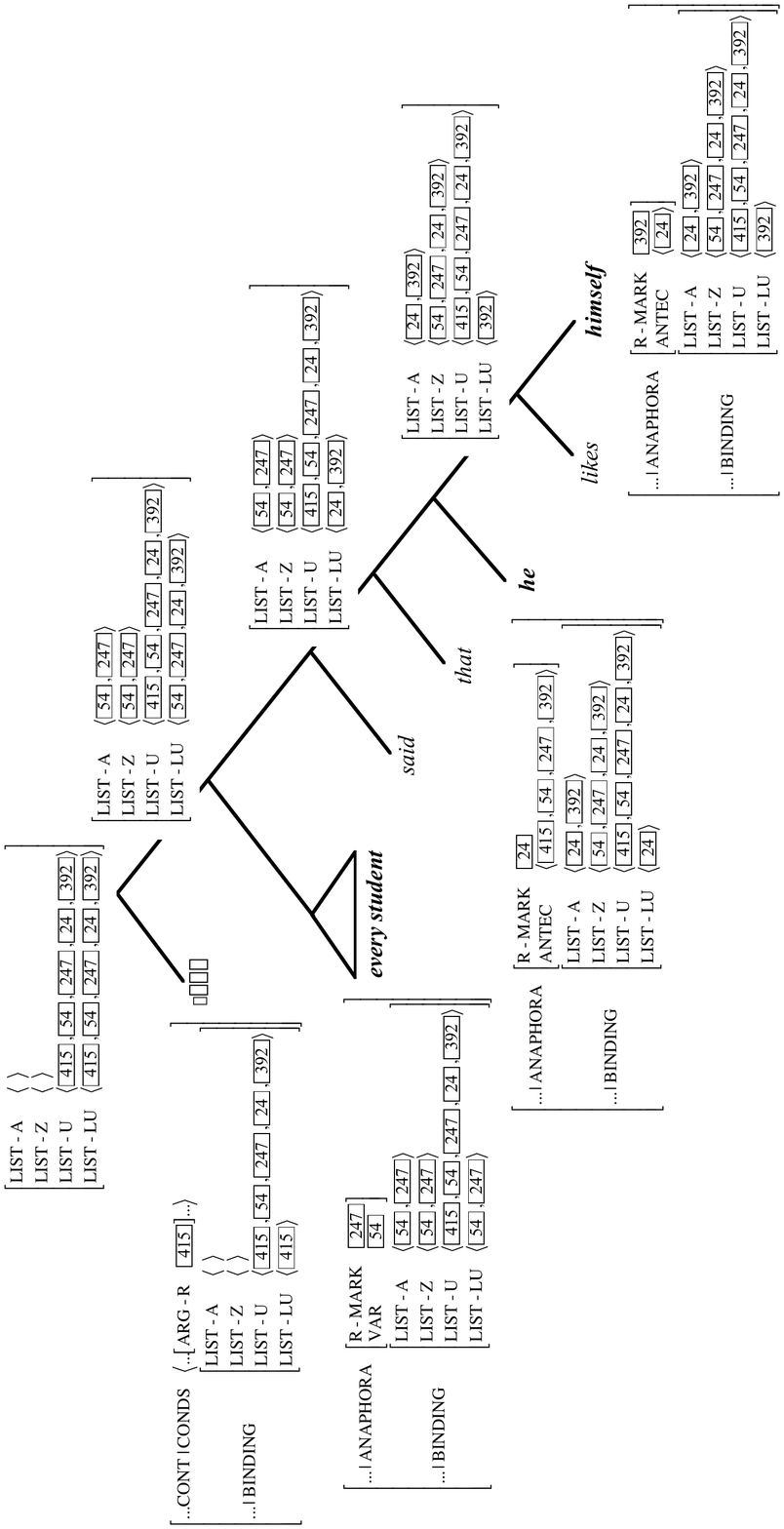} % jsilva
\caption{Partial grammatical representation of {\em Every student said he likes
himself}.}
\end{center}
\end{figure}

\subsection*{Circumscribing the Anaphoric Context}

Let us start by considering the representation of the context.

Taking
the representation of obliqueness hierarchies first, one can check that 
in the upper nodes of the  matrix clause, due to the effect of BDP--Clause~III, 
the {\sc list-z} value is obtained from the value of {\sc list-a},  with which it is
token-identical, thus comprising the list 
$\langle\raisebox{-0.7ex}{\begin{avm}\@{54},\@{247}\end{avm}}\rangle$.
In the nodes of the embedded clause, in turn, the {\sc list-z} value is the concatenation 
of that upper {\sc list-z} value and the {\sc list-a} value in the embedded clause
$\langle\raisebox{-0.7ex}{\em \begin{avm}\@{24}, \@{392}\end{avm}}\rangle$, 
from which the list 
$\langle\raisebox{-0.7ex}{\em \begin{avm}\@{54}, \@{247}, \@{24},
\@{392}\end{avm}}\rangle$  is obtained. 

In any point of the grammatical 
representation,
the {\sc list-a} values are obtained from the 
subcategorisation frames of the local verbal predicators, as constrained
by BDP--Clause II and the lexical constraints in (\ref{lexconst}) 
and (\ref{lexconst2}). Therefore,
$\langle\raisebox{-0.7ex}{\begin{avm}\@{24},\@{392}\end{avm}}\rangle$ 
is the {\sc list-a} value of {\em likes}, and 
$\langle\raisebox{-0.7ex}{\begin{avm}\@{54},\@{247}\end{avm}}\rangle$
is the {\sc list-a} value of {\em said}.

Taking into account {\sc list-lu}, as one ascends in 
the representation of the syntactic constituency, the list gets longer since, by the effect 
of BDP--Clause~I, the {\sc list-lu} value at a given node gathers the reference markers 
of the nodes dominated by it.  Consequently, at the discourse top
node, {\sc list-lu} ends up as a list including all reference markers: 
both those introduced in the discourse by the NPs in the example sentence
and \raisebox{-0.7ex}{\em \begin{avm}\@{415}\end{avm}}, the one
available in the non
linguistic context, from which the list 
$\langle\raisebox{-0.7ex}{\em \begin{avm}\@{415}, \@{54}, \@{247}, \@{24},
\@{392}\end{avm}}\rangle$ is the result. 

Note that in cases where the discourse contains
more than one sentence, BDP--ClauseI~(i.) ensures that
{\sc list-lu} ends up with  all reference markers 
from every sentence of the discourse. 

BDP--Clause I also ensures that
this list of all reference  markers is passed to the {\sc list-u} value of the
top node, and that this {\sc list-u} value is then percolated down to all nodes of the 
grammatical representation, including the nodes of anaphoric nominals.

\subsection*{Circumscribing the Anaphoric Potential}\label{circAnaphPotential}

To consider the representation of the NPs,
we should take a closer look at the leaf nodes in the constituency tree. 

\textbf{Contribution to the context} Let us
consider first how the NPs contribute to the representation of
the context.

Every phrase contributes 
to the global anaphoric context by passing 
the tag of its reference marker into its own {\sc list-lu}.  

In the case of a quantificational NP, like
\emph{every student}, two tags are passed, corresponding to the 
{\sc var} value \raisebox{-0.7ex}{\em \begin{avm}\@{54}\end{avm}} --- token-identical
with the {\sc dref} value of the restrictor and
providing for bound-variable anaphora interpretations --- and the
{\sc r-mark} value
\raisebox{-0.7ex}{\em \begin{avm}\@{247}\end{avm}} --- providing for
e-type anaphora. 
\is{e-type anaphora}

While the semantic types of anaphora --- including bound-variable and e-type anaphora --- are addressed in further detail in Section~\ref{semanticTypes},
it is of note at this point that a DRT account of e-type anaphora is followed here.\footnote{
\citep[p.311ff]{kamp:drt93}.} 
Accordingly, a quantificational NP contributes a plural reference marker to the
semantic representation of the discourse that may serve as the antecedent in (e-type) 
anaphoric links. In a sentence like {\em Every bald man snores}, for instance,
the quantificational NP contributes 
the plural reference marker which stands for the bald
men that snore. Such marker is introduced in the discourse representation
via the application of the DRT Abstraction operator \textSigma, which
takes the restrictor and the nuclear scope of the determiner and
introduces the plural marker that satisfies the corresponding
semantic conditions.\footnote{
\citep[p.310]{kamp:drt93}.}

In order to incorporate such an account of e-type anaphora
into Underspecified DRT,\footnote{
\citep{frank:sem95}.} 
the reference marker standing for the plurality
satisfying the semantic condition obtained with  
\mbox{\textSigma-abstraction}, in the {\sc conds} value of a determiner, 
is made token-identical with its {\sc r-mark} value. The {\em synsem} of the
lexical  entry for {\em every}, for instance, results thus as follows, where
\raisebox{-0.7ex}{\em \begin{avm}\@{1}\end{avm}} is the marker obtained
via 
\mbox{\textSigma-abstraction}:
%\footnote{
%Section \ref{reverse} below presents further discussion on 
%the capacity of quantificational NPs to act as antecedents
%in anaphoric links and on its constraints.
%}
% 
\\
\avmfont{\sc}
\avmvalfont{\it}
\avmsortfont{\it}
\begin{center}
\begin{avm}

\[loc\|cont & \[ ls & \[l-max & \@4 \\
                       l-min & \@5 \]\\
																	subord & \rm \{\@4>\@3,\@4>\@5,\@8$\geq$\@5\} \\
																	conds & \{\[label & \@4\\
                        					rel & every\\
																													res & \@3\\
																													scope & \@5 \],
																												\[label & \@3\\
                        					dref & \@2 \],\\
																												\[label & \@5\\
                        					dref & \@7 \],
																												\[label & \@8\\
                        					rel & \textSigma-abstraction\\
																													arg1 & \@2\\
																													arg2 & \@7\\
																													dref & \@1 \]
																										\}\\
             				anaph & \[r-mark & \@1\\
                          var & \@2 \] \] \\
  nonloc\|bind & \[list-a & list(refm) \\
																										list-z & list(refm) \\
																										list-u & list(refm) \\
																										list-lu & \<\@2, \@1\> \]  \]

\end{avm}
\end{center}

\textbf{Contribution by the context} Let us consider now at how the representation of the context
is encoded in each~NP.

It should be noted that the suitable values of {\sc list-a}, {\sc list-z} 
and {\sc list-u} at the different NP nodes are enforced 
by the combined effect of the three Clauses of BDP. 

Due to, respectively, BDP--Clause II (iii.) and BDP--Clause I (iii.i.), {\sc list-z} and 
{\sc list-u} values result from token-identity, respectively, with {\sc list-z}
and with {\sc list-u} of the immediately dominating node in the 
constituency tree --- that is the case, for instance, with the lists 
$\langle\raisebox{-0.7ex}{\begin{avm}\@{54},\@{247}\end{avm}}\rangle$
and 
$\langle\raisebox{-0.7ex}{\begin{avm}\@{415},\@{54},\@{247},\@{24},\@{392}\end{avm}}\rangle$
in the non-pronoun {\em every student} and in the sentential node dominating it. 

Due to BDP--Clause II (i.),
{\sc list-a} value, in turn, is obtained via token-identity with {\sc list-a} of
the subcategorising predicator --- that is the case, for instance, with the list 
$\langle\raisebox{-0.7ex}{\begin{avm}\@{24},\@{392}\end{avm}}\rangle$
in the reflexive {\em himself} and in its predicator {\em likes}.

\textbf{Realisation of anaphoric potential} As to the anaphoric nominals, let us consider how their
anaphoric potential is circumscribed in each specific occurrence.

The value of {\sc antec} is a
list that records the grammatically admissible antecedents of the
corresponding anaphor at the light of binding constraints.

As the result of the relational constraint {\em principleA}/2, the semantic
representation of the reflexive {\em himself} includes the attribute
{\sc antec} with the singleton list 
$\langle\raisebox{-0.7ex}{\em \begin{avm}\@{24}\end{avm}}\rangle$
as value, indicating that the only antecedent candidate
available in this sentence is the pronoun in the embedded clause whose
reference marker is identified as
\raisebox{-0.7ex}{\em \begin{avm}\@{24}\end{avm}} in its own
semantic representation. 

The semantic representation of the pronoun {\em he}, in turn,
includes the feature \mbox{{\sc antec}} with a value that is the list
of its antecedent candidates, 
$\langle\raisebox{-0.7ex}{\em \begin{avm}\@{415}, \@{247}, \@{54},
\@{392}\end{avm}}\rangle$, thus indicating that, in this sentence,
the pronoun can be anaphorically linked to every nominal except itself, 
in line with the relational constraint {\em principleB}/3. 

This {\sc antec} list includes antecedent
candidates for the pronoun that will be dropped out by preferences
or constraints on
anaphoric links other than just the grammatical binding constraint expressed in Principle B. 
For instance,
the plural reference marker \raisebox{-0.7ex}{\em \begin{avm}\@{247}\end{avm}},
which is the {\sc r-mark} value of {\em every student},
will eventually be excluded by the anaphora resolver
given that the singular pronoun {\em he} cannot
entertain an e-type anaphoric link with a universally quantified NP
whose reference marker obtained by 
\mbox{\textSigma-abstraction} is a plurality.

Also the marker \raisebox{-0.7ex}{\em \begin{avm}\@{392}\end{avm}}
of the reflexive
will be eventually discarded from this {\sc antec} list as a suitable antecedent by the resolver system
since this would lead to an interpretive loop where the pronoun and the
reflexive would be the sole antecedents of each other.

\textbf{Non-linguistic context} Finally, in order to illustrate how the non linguistic context may be represented 
in the linguistic representation of sentences, 
in this example, the reference marker  
\raisebox{-0.7ex}{\em \begin{avm}\@{415}\end{avm}}
was introduced in the semantic representation of the \texttt{ctx} node. 

The {\sc conds} value of this node is meant to capture the possible contribution 
of the non-linguistic context at stake for the interpretation of the discourse. 
Like in the lexical entries of nominals, in the feature representation of {\em ctx},
the reference marker 
\raisebox{-0.7ex}{\em \begin{avm}\@{415}\end{avm}}
is integrated in the {\sc list-lu} value.
By the effect of BDP--Clause I, this reference marker ends up added to the list of 
all reference markers, from both the linguistic discourse and the non-linguistic
context, which is the shared value of features {\sc list-lu} and {\sc list-u} at the top node in Figure 1.

\section*{Abbreviations}

BDP - Binding Domains Principle\\
DRT - Discourse Representation Theory\\
HPSG - Head-Driven Phrase Structure Grammar\\
UDRT - Underspecified Discourse Representation Theory

\section*{Acknowledgements}

The research reported in this paper was partially supported by the 
Research Infrastructure  for  the Science  and  Technology of Language (\mbox{PORTULAN CLARIN}).

{\sloppy
\printbibliography[heading=subbibliography,notkeyword=this]
}
\end{document}